\documentclass[10pt,twocolumn,letterpaper]{article}
\pdfoutput=1
\usepackage[accsupp]{axessibility}  
\usepackage{iccv}
\usepackage{times}
\usepackage{epsfig}
\usepackage{graphicx}
\usepackage{amsmath}
\usepackage{amssymb}
\usepackage{capt-of}
\usepackage{microtype}
\usepackage{graphicx}
\usepackage{subfigure}
\usepackage{booktabs} 
\usepackage{multirow}
\usepackage{colortbl}  
\usepackage{xcolor}
\usepackage{array} 
\usepackage{float}
\usepackage[export]{adjustbox}
\usepackage[pagebackref=true,breaklinks=true,letterpaper=true,colorlinks,bookmarks=false]{hyperref}

\usepackage{threeparttable}
\usepackage[ruled,linesnumbered]{algorithm2e}
\usepackage{amsmath}
\usepackage{amssymb}
\usepackage{mathtools}
\usepackage{amsthm}

\def\mF{{\mathcal F}}

\def\mL{{\mathcal L}}

\def\mN{{\mathcal N}}

\def\mS{{\mathcal S}}
\def\mT{{\mathcal T}}

\def\0{{\bf 0}}
\def\1{{\bf 1}}

\def\bP{{\bf P}}

\def\bT{{\bf T}}

\def\bc{{\bf c}}
\def\bd{{\bf d}}

\def\bk{{\bf k}}

\def\bo{{\bf o}}

\def\br{{\bf r}}

\def\bx{{\bf x}}

\def\mmE{{\mathbb E}}

\def\bx{{\bf x}}

\def\bP{{\bf P}}

\def\bc{{\bf c}}

\newtheorem{prop}{Proposition}

\newcommand{\secondbest}{\underline}

\def\eg{\emph{e.g.,}} 
\def\ie{\emph{i.e.,}} 
\def\cf{\emph{c.f.}} 
\def\etc{\emph{etc.}} 
\def\wrt{{w.r.t.}} 
\def\etal{\emph{et al.~}}
\def\sexyname{CR-NeRF}

\usepackage{algpseudocode}
\usepackage{forloop}

\theoremstyle{plain}

\theoremstyle{definition}

\theoremstyle{remark}

\usepackage[breaklinks=true,bookmarks=false]{hyperref}
\usepackage[capitalize,noabbrev]{cleveref}

\iccvfinalcopy 

\def\iccvPaperID{2468} 

\ificcvfinal\fi

\begin{document}
\def\mytitle{Cross-Ray Neural Radiance Fields for Novel-view Synthesis \\from Unconstrained Image Collections}
\title{\mytitle}
\author{Yifan Yang$^{1,2*}$~~Shuhai Zhang$^{1,2}$~~Zixiong Huang$^1$~~Yubing Zhang$^3$~~~Mingkui Tan$^{1,2,4}$\footnotemark[2]\\
$^1$South China University of Technology~~$^2$Pazhou Lab~~$^3$Guangzhou Shiyuan Electronics Co., Ltd\\
$^4$Key Laboratory of Big Data and Intelligent Robot, Ministry of Education\\
{\tt\small \{seyoungyif, mszhangshuhai, sesmilhzx\}@mail.scut.edu.cn, zhangyubing@cvte.com} \\ {\tt\small mingkuitan@scut.edu.cn}
}
\maketitle
\ificcvfinal\thispagestyle{empty}\fi

\renewcommand{\thefootnote}{\fnsymbol{footnote}}
\footnotetext[2]{Corresponding author.}
\footnote{This work was done when Yifan Yang was a
research intern at Guangzhou Shiyuan Electronics Co., Ltd.}
\renewcommand{\thefootnote}{\arabic{footnote}}

\begin{abstract} 
   Neural Radiance Fields (NeRF) is a revolutionary approach for rendering scenes by sampling a single ray per pixel and it has demonstrated impressive capabilities in novel-view synthesis from static scene images. However, in practice, we usually need to recover NeRF from unconstrained image collections, which poses two challenges: 1) the images often have dynamic changes in appearance because of  different capturing time and camera settings;  2) the images may contain transient objects such as humans and cars, leading to occlusion and ghosting artifacts. Conventional approaches seek to address these challenges by \textit{locally} utilizing a \textbf{single ray} to synthesize a color of a pixel. In contrast, humans typically perceive appearance and objects by globally utilizing information across multiple pixels. To mimic the perception process of humans, in this paper, we propose Cross-Ray NeRF (CR-NeRF) that leverages interactive information across \textbf{multiple rays} to synthesize occlusion-free novel views with the same appearances as the images. Specifically, to model varying appearances, we first propose to represent multiple rays with a novel cross-ray feature and then recover the appearance by fusing global statistics, i.e., feature covariance of the rays and the image appearance.  Moreover, to avoid occlusion introduced by transient objects, we propose a transient objects handler and introduce a grid sampling strategy for masking out the transient objects. We theoretically find that leveraging correlation across \textit{multiple rays} promotes capturing more \textit{global information}. Moreover, extensive experimental results on large real-world datasets verify the effectiveness of CR-NeRF. 
\end{abstract}

\section{Introduction}

\label{Introduction}
Novel-view synthesis is a long-standing problem in computer vision that has paved the way for numerous applications such as virtual reality and digital humans~\cite{guo2021ad,su2021nerf}. More recently, the emergence of Neural Radiance Fields (NeRF) has driven the field forward, as it has shown significant performance in reconstructing 3D geometry~\cite{wang2021neus} and recovering the appearance~\cite{chen2021mvsnerf, park2021nerfies, barron2022mip} from multi-view image sets. However, NeRF assumes that the images do not have variable appearances and moving objects~\cite{martinbrualla2020nerfw} (called the \textit {static scene assumption}~\cf~Sec.~\ref{sec:back_nerf}), which leads to significant performance degradation on large-scale Internet image collections. To expand the scope of NeRF, we aim to exploit the collections and provide a 3D immersive experience through which we can visit international landmarks such as the Brandenburg Gate, and the Trevi Fountain from different viewpoints and times of one day.

\par To achieve this, we address the problem of recovering an appearance-controllable and anti-occlusion NeRF from unconstrained image collections. In other words, by reconstructing the NeRF representation, we control the appearance of the scene based on photos with various photometric conditions, while eliminating occlusions caused by the images. Although providing a sense of immersion, reconstructing NeRF with these images faces the following two challenges. 1) \textit{Varying appearances}: Imaging two tourists who take photos in the same viewpoint but under various conditions, \eg~different capturing times, diverse weather (\eg~sunny, rainy, and foggy), and different camera settings (\eg~aperture, shutter, and ISO). This varying condition causes that although multiple photographs are taken of the same scene, they look dramatically different. 2) \textit{Transient occlusion}: Even with a constant appearance, transient objects such as cars and Pedestrians may obscure the scene. Since these objects are usually captured by only one photographer, it is usually impractical to reconstruct these objects in high quality. The above challenges conflict with the static-scene assumption of NeRF and result in inaccurate reconstruction that leads to over-smoothing and ghosting artifacts~\cite{martinbrualla2020nerfw}.
\begin{figure}[t]
    \centering
\includegraphics[width=0.5\textwidth]{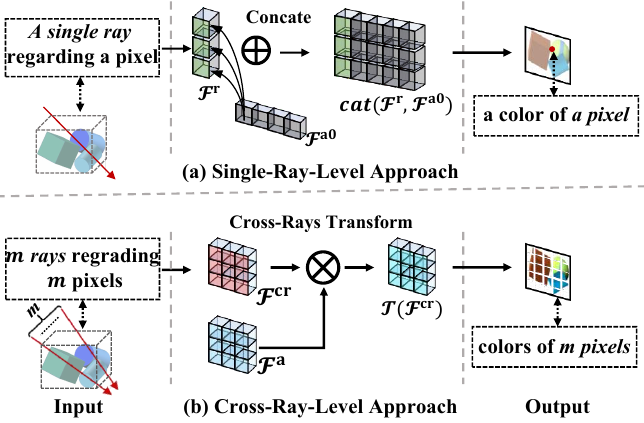}
    \caption{Illustration of single-ray-level and cross-ray-level approaches. The conventional one (a) generates each pixel regarding a \textit{single ray} independently. In contrast, our proposed \sexyname~(b) considers information of \textit{multiple rays} and synthesizes a patch simultaneously. $\mathcal{F}^{a}$ and $\mathcal{F}^{a_0}$ are conditioned features.}
    \label{fig:movivation}
\end{figure}
\par Recently, several attempts (NeRF-W~\cite{martinbrualla2020nerfw};Ha-NeRF~\cite{chen2022hallucinated}) have been proposed to address the aforementioned challenges.  From  Fig.~\ref{fig:movivation}(a), NeRF-W and Ha-NeRF leverage a single-ray manner, wherein \textit{a single camera ray} (\ie~a beam of light extending from a camera through a pixel on an image plane into a 3D scene) serves as input. This manner then involves considering appearance and occlusion factors and subsequently synthesizing \textit{each color of pixel } of a novel view independently. One potential issue of this manner is its reliance on local information (\eg~information of a single image pixel) of every single ray for recognizing appearance and transient objects. In contrast, humans tend to utilize global information (\eg~information across multiple image pixels), which provides a more comprehensive understanding of an object to observe its appearance and handle occlusion. 
Motivated by this,
we propose to tackle varying appearance and transient objects with a cross-ray paradigm (see Fig.~\ref{fig:movivation}(b)), wherein we utilize global information
from 
multiple rays to recover the appearance and handle transient objects. Subsequently, we synthesize a region of a novel view simultaneously.  Based on the cross-ray paradigm, we propose a Cross-Ray Neural Radiance Fields (\sexyname), {which comprises two components:} 1) To model variable appearances, we propose to represent information of multiple rays with a novel cross-ray feature. We then fuse the cross-ray feature and an appearance embedding 
via a cross-ray transformation network using global statistics, \eg~feature covariance of the cross-ray.
The fused feature is fed to a decoder to obtain colors of several pixels simultaneously. 2) {To handle transient objects,} we propose a unique perspective of handling transient objects as a segmentation problem, through which we detect transient objects by considering global information of an image region. From this perspective, we segment the unconstrained images for a visibility map of the objects. To avoid computation overhead, we introduce a grid sample strategy that samples the segmented maps to pair with the input rays.  We theoretically analyze that leveraging correlation across multiple rays promotes capturing more global information. 
\par We summarize our contributions in three folds:
    \par $\bullet$ A new cross-ray paradigm for novel-view synthesis from unconstrained photo collections: We find that existing methods fall short of producing satisfactory visual outcomes from unconstrained photo collections via a single-ray-level paradigm, primarily due to the neglect of the potential cooperative interaction among multiple rays. To address this, we propose a novel cross-ray paradigm, which exploits the global information across multiple rays.
    
    \par $\bullet$ An {interactive and global scheme} for addressing varying appearances: Unlike existing methods that process each ray independently, we represent multiple rays by introducing a cross-ray feature, which facilitates the interaction among rays through feature covariance. This enables us to inject a global informative appearance representation into the scene, resulting in more realistic and efficient appearance modeling. Our theoretical analysis demonstrates the necessity of considering multiple rays for appearance modeling.
     \par $\bullet$ A novel segmentation technique for processing transient objects: We reformulate the transient object problem as a segmentation problem. We use global information of an unconstrained image to segment a visibility map. Moreover, we apply grid sampling to pair the map with multiple rays. Empirical results show that \sexyname~eliminates the transient objects in reconstructed images.  

\begin{figure*}[th]
    \centering    \includegraphics[width=1\textwidth]{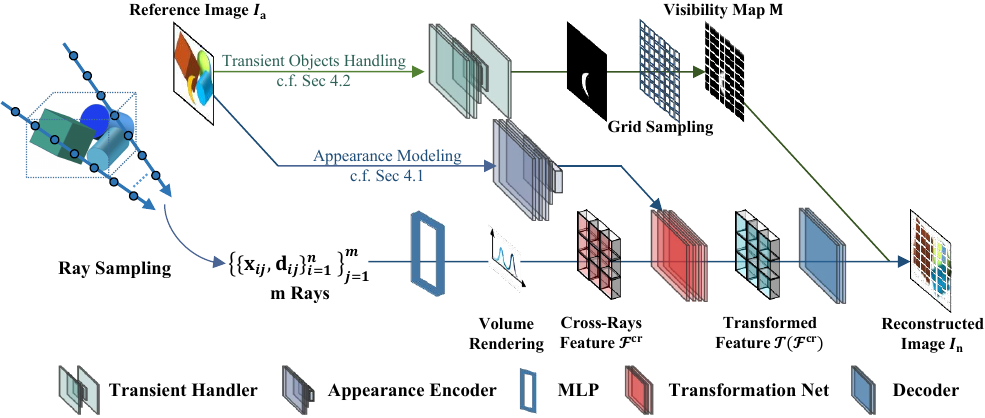}
    \caption{An overview of Cross-Ray Neural Radiance Field (\sexyname). Given position $\textbf{x}$ and direction $\textbf{d}$ of multiple rays, We first generate a cross-ray feature $\mathcal{F}^{\mathrm{cr}}$ that accumulates multi-view information in a scene. To incorporate the appearance information of the reference image $\mathcal{I}_{a}$, the appearance encoder is used to learn the appearance features $\mathcal{F}_{a}$, the transform net to fuse $\mathcal{F}_{a}$ with $\mathcal{F}^{\mathrm{cr}}$ and the decoder to synthesize the colors of multiple pixels in the reconstructed image $\mathcal{I}_{\mathrm{n}}$ simultaneously. To eliminate transient objects in $\mathcal{I}_{a}$,  our transient handler generates a visibility map, for which we introduce a grid sampling strategy to match the map with the rays during training. }
    \label{fig:overview}
\end{figure*} 
\section{Related Works}
\par \textbf{Neural rendering.}
Neural rendering applies deep learning with computer graphic technologies to render images and reconstruct 3D scenes. Recent advances seek to apply learning-based technology to generate representations such as signed distance field ~\cite{park2019deepsdf,liu2020dist,zhang2021learning, jiang2020sdfdiff}, point clouds~\cite{fan2017point,jiang2018gal,lin2018learning}, voxels~\cite{peng2022tmvnet, henzler2019escaping, xie2019pix2vox} and occupancy fields~\cite{xu2021generative, mescheder2019occupancy, Peng2020ECCV}, which are then applied for rendering novel views. With the remarkable performance, NeRF~\cite{mildenhall2021nerf} has attracted attention from the neural rendering community. More recently, NeRF has been extended to represent a time-series of scenes~\cite{li2021neural, pumarola2021d, li2022neural}, handle high-resolution settings~\cite{jiang2023alignerf, wang20224k}, address relighting~\cite{srinivasan2021nerv}, and reconstruct large-scale environments~\cite{tancik2022block, turki2022mega,turki2023suds}. 
Notably, one limitation of NeRF is that it assumes the scene is static, which faces challenges of varying appearance and presence of transient objects in unconstrained image collections. To alleviate this, NeRF-W~\cite{martinbrualla2020nerfw} and Ha-NeRF~\cite{chen2022hallucinated} focus on addressing the challenges by processing each ray of a scene independently. Differently, we propose to accumulate information on multiple rays for modeling appearance and eliminating transient objects.

\par \textbf{Novel-view synthesis.}
Synthesizing views from a novel viewpoint has long been a fundamental problem in computer vision and computer graphics. Traditionally, novel views can be synthesized through 4D light field strategy~\cite{levoy1996light, Wilburn2005High, chen1993view}. However, the strategy requires a dense camera array for capturing data, which is usually impractical. Since collecting a sparse set of images are efficient, view synthesis research takes advantage of geometry structure~\cite{buehler2001unstructured,cheng2008improved} to aid in constructing novel views with limited input. With the flourish of deep learning, deep neural networks have been leveraged to estimate the scene geometry (\eg~point clouds~\cite{wiles2020synsin,rockwell2021pixelsynth}, depth map~\cite{srinivasan2017learning,lo2018food}, multiple-layer image~\cite{flynn2019deepview,wizadwongsa2021nex}). Although leveraging the geometry enhances the quality of novel views, the estimation is usually without ground truth supervision and usually is not accurate enough. To circumvent the difficulty of estimating precise geometry, we propose to utilize an implicit function, \ie~neural radiance fields (NeRF)~\cite{mildenhall2021nerf} for novel-view synthesis.

\section{Preliminaries}
\label{sec:back_nerf}
Neural Radiance Fields (NeRF)~\cite{mildenhall2021nerf} implicitly represents a static 3D scene with \textit{multilayer perceptron} (MLP) and then produces a novel view via \textit{volume rendering} (VR)~\cite{drebin1988volume}. 
NeRF generates a pixel color of a novel view from a camera ray independently. In this sense, we can describe the rendering process \wrt~a single camera ray $\br(t)=\bo+t \bd$ which is cast from a camera center $\bo$ in the direction $\bd$ and passes through a pixel on an image plane \wrt~the novel view.  We sample n ray points $\{\br(t_{i})\}_{i=1}^{n}$ along $\br$ between a given near plane $t_n$ and a far plane $t_f$. For each ray point $\br(t_i)$, we query the MLP at a 3D position $\bx_i=(x,y,z)$ and a viewing orientation $\bd_i=(d_x,d_y,d_z)$ to obtain a color $\bk_i=(r,g,b)$ and a density $\mathbf{\sigma}_i$ via equations:
               $\bx_i \to  \{\mathcal{F}^{r}_{i}, \mathbf{\sigma}_{i}\}$,
                $\{\mathcal{F}^{r}_{i}\, \bd_i\} \to \bk_i$,
where $\mathcal{F}^{r}_{i}$ denotes a ray-point-level feature regarding the ray point $i$
. To learn high-frequency information, position encoding~\cite{mildenhall2021nerf} is employed to $\bx_i$ and $\bd_i$.  Typically, $\bo$ and $\bd$ are estimated by structure from motion approaches~\cite{schonberger2016structure, ullman1979interpretation} from multi-view images regarding the 3D scene.
\par To approximate the color $\hat{\bc}(\br)$ of the pixel of a reference image, NeRF accumulates n ray points $\{\br(t_{i})\}_{i=1}^{n}$ along the ray $\br$ into the $\hat{\bc}(\br)$ via VR~\cite{drebin1988volume}:
\begin{equation}
\label{eq:nerf2}
\begin{aligned}
\hat{\bc}(\br) =\sum_{i=1}^{n} \varphi_{i}\mathbf{\alpha}_i \bk_{i}, \varphi_{i} =\exp (&-\sum_{l=1}^{i-1} \mathbf{\sigma}_{l} \mathbf{\delta}_{l}),\\~\mathbf{\alpha}_i = 1-\exp \left(-\mathbf{\sigma}_{i} \mathbf{\delta}_{i}\right).
\end{aligned}
\end{equation}
Here, $\mathbf{\alpha}_i$ is the probability of the ray that terminates at $\br(t_{i})$;  $\varphi_i$ is the accumulated transmittance from the near plane $t_n$ to $\br(t_{i})$; $\mathbf{\delta}_{l}=t_{l+1}-t_{l}$ is distance between two adjacent ray points. The MLP is optimized via minimizing the loss:
$\mathcal{L}=||\hat{\bc}(\br)- \bc(\br)||_2^2$,
where $\bc(\br)$ denotes the ground truth color of a pixel \wrt~the ray $\br$.
\par \textbf{Limitations of NeRF on novel-view synthesis from unconstrained collections.} 
    Given an unconstrained collection of a scene, we seek to reconstruct the scene whose appearance can be modified according to a new image, while removing transient objects. Since NeRF assumes the lighting in the scene is constant over time and there are no moving objects or changes in lighting during the time that the input images are captured (called \textit{static scene assumption}~\cite{martinbrualla2020nerfw}),  NeRF is limited to effectively modeling the geometry and appearance of static scenes only. 
    To address the limitation of NeRF, recent advances~\cite{martinbrualla2020nerfw, chen2022hallucinated} synthesize novel views on single-ray level (see Fig.~\ref{fig:movivation} (a)) following equation:
\begin{equation}
\label{eq:style_previous}
\begin{aligned}
  \{\bx_i,\bd_i, \mathcal{F}^{a_0}\}_{i=1}^{n} \to \hat{\bc}_{\mathrm{n}},
  \end{aligned}
\end{equation}
where $\mathcal{F}^{a_0}$ is image-level conditional embedding of $\mathcal{I}_{a}$. {From Eqn.~\ref{eq:style_previous}, existing methods~\cite{martinbrualla2020nerfw, chen2022hallucinated} generate the color $\hat{\bc}_{\mathrm{n}}$ of each pixel by processing the corresponding single ray independently. This manner}
ignores global information among multiple rays, leading to inaccurate appearance modeling (see our empirical studies in Fig.~\ref{fig:comparison_all} and Fig.~\ref{fig:transfer_style}).

\section{Cross-Ray Neural Radiance Fields}
Given unconstrained photo collections of a scene, we seek to reconstruct the scene whose appearance can be modified based on a new image, while removing transient objects. 
This task is challenging due to the existence of variable appearances and transient occlusions in the photo collections. To address this, intuited by that a human usually detects appearance and transient objects by considering global information (\eg~information across several image pixels) rather than local information (\eg~information of a single image pixel), we propose a Cross-Ray Neural Radiance Fields (\sexyname) that exploits global information across multiple camera rays, which correspond to several pixels of an image, to address both challenges.

\par  As shown in Fig.~\ref{fig:overview} and Alg.~\ref{alg:crnerf}, \sexyname~consists of two components: 1) \textit{Cross ray appearance modeling}~(\cf~Sec.~\ref{sec:appearance}). To model varying appearances, we first sample a grid of rays using a grid sampling strategy~\cite{Schwarz2020NEURIPS}. Next, we represent the rays with a novel cross-ray feature $\mathcal{F}^{\mathrm{cr}}$. 
We then inject an appearance embedding $\mathcal{F}^{a}$ into $\mathcal{F}^{\mathrm{cr}}$ via a learned transformation network. The fused feature is fed to a decoder for obtaining colors of multiple pixels simultaneously. We theoretically analyze the necessity of considering multiple rays and thus design an appearance loss $\mathcal{L}_a$ for cross-ray appearance modeling. 2) \textit{Cross-ray transient objects handling} (\cf~Sec.~\ref{sec:occlusion}). To handle transient objects, we deploy a segmentation network for generating a visibility map regarding transient objects.  To pair the map with the rays, we also apply the grid sampling strategy on the maps. We devise an occlusion loss $\mathcal{L}_t$ for transient handling. 

The overall optimization of our proposed \sexyname~minimizes the following objective function: 
\begin{equation}
    \mathcal{L}_{overall} = \mathcal{L}_{a}+\lambda \mathcal{L}_{t}, 
\end{equation}
where $\lambda$ is a hyper-parameter for balancing the appearance loss $\mathcal{L}_{a}$ (see Eqn.~\ref{loss: transfer}) and the occlusion loss $\mathcal{L}_{t}$ (see Eqn.~\ref{eqn:generate_mask}).

\subsection{Cross-Ray Appearance Modeling}
\label{sec:appearance}
To adapt \sexyname~to variable appearance through a global perspective, we modify the scene by leveraging multiple rays and the appearance of the unconstrained images.
\par \textbf{Representing scene information with multiple rays.} To model appearance from a multi-view observation, we first represent scene information using multiple rays. To this end, we propose a novel \textit{cross-ray feature} $\mathcal{F}^{\mathrm{cr}}$ with equations:
\begin{equation}
\begin{aligned}
  \{\{\mathcal{F}^r_{ij}, \sigma_{ij}\}_{i=1}^{n}\}_{j=1}^{m} = &\{\mathrm{MLP}_{\theta_1}(\{\bx_{ij},\bd_{ij}\}_{i=1}^{n})\}_{j=1}^{m},\\
  \mathcal{F}^{\mathrm{cr}}= \{\mathrm{VR}(\{&\mathcal{F}^r_{ij}, \sigma_{ij},\delta_{ij}\}_{i=1}^{n})\}_{j=1}^{m},
  \end{aligned}
\label{eq:style_ours_generating}
\end{equation}

Besides, we obtain an appearance feature $\mathcal{F}^{a}$ of an appearance image $\mathcal{I}_{a}$ by $\mathcal{F}^{a}=E_{\theta_2}(\mathcal{I}_{a})$.
With $\mathcal{F}^{\mathrm{cr}}$ and $\mathcal{F}^{a}$, it is critical to find an effective fusion manner to inject image appearance into the scene representation.

\par  \textbf{Injecting appearance into scene representation.} {The key of our cross-ray appearance modeling is to exploit the potential cooperative relationship among the cross-ray features $\mathcal{F}^{\mathrm{cr}}$ to facilitate appearance modeling from the given appearance image $\mathcal{I}_{a}$ to the scene representation.  In other words, we seek a transformation operation that can transfer the style from a reference image and also retain the essential content during training. To this end, we learn a transformation $\mT$ to align the transferred cross-ray features $\mT(\mF^{\mathrm{cr}})$ and the appearance feature $\mF^a$ with an auxiliary identity term, which is formulated as below,     
\begin{equation}
\label{Prob: appearance modeling}
\begin{aligned}
    \min_\mT \mmE_{\mF^{\mathrm{cr}}, \mF^a} \|\mT(\mF^{\mathrm{cr}}) {-}\mF^a\|_{2}^{2}+\beta \|\bP\mT(\mF^{\mathrm{cr}}){-} \mF^{\mathrm{cr}}\|_{2}^{2},
\end{aligned}
\end{equation}
where $\beta$ is a trade-off parameter and $\bP$ is a constant matrix for matching the transformed feature $\mT(\mF^{\mathrm{cr}})$ and $\mF^{\mathrm{cr}}$. 
} 
Next, we theoretically analyze the necessity of considering multiple rays to solve Problem (\ref{Prob: appearance modeling}) for appearance modeling.

 \begin{algorithm}[t]
 \small 
	\caption{The training pipeline of \sexyname.}
	\label{alg:crnerf}
	\KwIn{$m$ rays, a reference image $\mathcal{I}_{a}$, a multilayer perceptron $\mathrm{MLP}_{\theta_1}$, an appearance encoder $E_{\theta_2}$, a transformation net $\mT_{\theta_3}$, a decoder $D_{\theta_4}$, a content encoder $E_{\theta_5}$ and a segmentation net $\mS_{\theta_{\Delta}}$.}
 \KwOut{The estimated colors of $m$ pixels of a novel view.}

	\While{\textnormal{not converged}}{
		Generate cross-ray features $\mF^{\mathrm{cr}}$
and appearance feature $\mF^{a}$ with $E_{\theta_2}$ and $\mathrm{MLP}_{\theta_1}$ by Eqn. (\ref{eq:style_ours_generating}).

    Obtain the loss $\mL_a$ for modeling appearance with $E_{\theta_2}$, $\mT_{\theta_3}$, $D_{\theta_4}$ and $E_{\theta_5}$ by Eqn. (\ref{loss: transfer}).  \Comment{\cf~Sec.~\ref{sec:appearance}}
    
    Obtain the visibility map $\textbf{M}$ for masking out transient objects with and $\mS_{\theta_{\Delta}}$ by Eqn. (\ref{eqn:generate_mask}). 
    
    Obtain the loss $\mL_t$ for handling transient with $\mT_{\theta_3}$, $D_{\theta_4}$ and $\mS_{\theta_{\Delta}}$ by Eqn. (\ref{loss: transient}). \Comment{\cf~Sec.~\ref{sec:occlusion}}
    
    Obtain the overall loss of $\mL_{overall}=\mL_a +\lambda \mL_t$.
    
    Update the parameters $\Theta = \{\theta_1, \theta_2, \theta_3, \theta_4, \theta_5, \theta_{\Delta} \}$ by descending the gradient: $\nabla_{\Theta} \mL_{overall}$}
\end{algorithm}

\par \textbf{Necessity of considering multiple rays for appearance modeling.} We consider a Gaussian case that can provide some insights to devise an effective approach to inject the appearance into the scene representation. To this end, we assume the two features $\mF^a$ and $\mF^{\mathrm{cr}}$ are following two Gaussian distributions and $\mT$ is a linear transformation that rigorously matches two distributions.
We provide a closed-form solution to Problem (\ref{Prob: appearance modeling}) under this assumption as follows.
\begin{prop}
\label{thm: transfer matrix}
Given an invertible constant matrix $\bP {\in} \mathbb{R} ^{C {\times} C}$, assuming that $\mF^a {\sim} \mN(\boldsymbol{\mu}_a, \boldsymbol{\Sigma}_a)$,  $\mF^{\mathrm{cr}} {\sim} \mN(\boldsymbol{\mu}_{\mathrm{cr}}, \boldsymbol{\Sigma}_{\mathrm{cr}})$ and $\mT(\mF^{\mathrm{cr}}) {\sim} \mN(\boldsymbol{\mu}_a, \boldsymbol{\Sigma}_a)$, where $\mT(\mF^{\mathrm{cr}}) {=} \bT(\mF^{\mathrm{cr}} {-} \boldsymbol{\mu}_{\mathrm{cr}}) {+} \boldsymbol{\mu}_a$ and $\bT {\in} \mathbb{R}^{C {\times} C}$ is a transformation matrix, the optimal $\bT$ to Problem~\ref{Prob: appearance modeling} is:
\begin{equation}
\begin{aligned}
    \bT=\bP^{-1}\boldsymbol{\Sigma}_{\mathrm{cr}}^{-1 / 2}\left(\boldsymbol{\Sigma}_{\mathrm{cr}}^{1 / 2} \bP ^ \top \boldsymbol{\Sigma}_{a} \bP \boldsymbol{\Sigma}_{\mathrm{cr}}^{1 / 2}\right)^{1 / 2} \boldsymbol{\Sigma}_{\mathrm{cr}}^{-1 / 2}.
\end{aligned}
\end{equation}
\end{prop}
Proposition \ref{thm: transfer matrix} suggests the transformation matrix $\bT$ is determined by the covariance of $\mF^{\mathrm{cr}}$ and $\mF^a$ given $\bP$, which is consistent with the conclusion in~\cite{li2019learning, lu2019closed}. Inspired by this, we can construct a neural network to learn the appearance transformation $\mT$ by feeding the covariances of $\mF^a$ and  $\mF^{\mathrm{cr}}$.  Specifically, we adopt an effective transformation network following Li \etal~\cite{li2019learning} which is defined as: 
\begin{equation}
\begin{aligned}
    \mT(\mF^{\mathrm{cr}}) &= \bT  \hat{\mF}^{\mathrm{cr}},\\
    \mathrm{where}~\bT = Cov(&\bar{\mF^{\mathrm{cr}}})   Cov(\bar{\mF}^{a})),\\    
    \hat{\mF}^{\mathrm{cr}} = \phi_1(\mF^{\mathrm{cr}}),
    \bar{\mF}^{\mathrm{cr}}& = \phi_2(\mF^{\mathrm{cr}}), 
    \bar{\mF}^{a} = \phi_3(\mF^{a}).
\end{aligned}
\end{equation}

Here, $\phi_1$, $\phi_2$, and $\phi_3$ are non-linear mappings parameterized by convolutional neural networks (CNN) that can express richer embedding to prepare for appearance modeling.
Intuitively, we consider multiple rays when modeling appearance to employ multi-view information. The correlation between the feature maps of these different views, which can be given by the covariance, is able to capture more global texture information for a given appearance image~\cite{gatys2016image, li2017universal, dai2023disentangling}, thus facilitating better appearance modeling for a scene. 

\textbf{Loss function $\mL_a$ for varying appearance modeling:} To generate a novel-view image with a satisfactory appearance from the transformed feature $\mT_{\theta_3}(\mF^{\mathrm{cr}}))$, we need to enforce a decoder $D_{\theta_4}$ into the training process of appearance modeling. Inspired by the formulation in Problem (\ref{Prob: appearance modeling}),  we provide the loss function for appearance modeling as:
\begin{equation}
\begin{aligned}
\label{loss: transfer}
     \mathcal{L}_{a} &= ||E_{\theta_2}[D_{\theta_4}(\mT_{\theta_3}(\mF^{\mathrm{cr}}))]- \mF^{\mathrm{a}}||_2^{2}\\
    &+\beta ||E_{\theta_5}[D_{\theta_4}(\mT_{\theta_3}(\mF^{\mathrm{cr}}))]-E_{\theta_5}[D_{\theta_4}(\mF^{\mathrm{cr}})]||_2^{2}, 
\end{aligned}
\end{equation}
where $\mF^{\mathrm{cr}}$ is 
obtained with an $\mathrm{MLP}_{\theta_1}$ by Eqn.~\ref{eq:style_ours_generating}. Here, we use a tailored encoder $E_{\theta_5}$ to model the transformed feature $\bP \mT(\mF^{cr})$ so that the content of the transformed image closely matches its original counterpart. 
In this way, we can synthesize a novel-view image by $\mathcal{I}_{\mathrm{n}} = D_{\theta_4}(\mT_{\theta_3}(\mF^{\mathrm{cr}}))$.

\subsection{Transient Objects Handling}
\label{sec:occlusion}

\begin{table*}[th]
\centering
 \begin{center}
 \begin{threeparttable}
 \large
\setlength{\tabcolsep}{0pt} 
\resizebox{1\linewidth}{!}{
 	\begin{tabular}{c|l|ccccccccc}
\toprule
\multicolumn{2}{c}{\multirow{2}{*}{}} & \multicolumn{3}{c}{Brandenburg Gate} & \multicolumn{3}{c}{Sacre Coeur} & \multicolumn{3}{c}{Trevi Fountain} \\
\cmidrule(lr){3-5} \cmidrule(lr){6-8} \cmidrule(lr){9-11} 
\multicolumn{2}{c}{} &~~PSNR ($\uparrow$) &~~SSIM ($\uparrow$) &~~LPIPS ($\downarrow$) &~~PSNR ($\uparrow$) &~~SSIM ($\uparrow$) &~~LPIPS ($\downarrow$) &~~PSNR ($\uparrow$) &~~SSIM ($\uparrow$) &~~LPIPS ($\downarrow$) \\
\midrule
    \multicolumn{1}{c|}{\multirow{5}[2]{*}{~~\rotatebox{90}{R / 4}~~}} &~NeRF & 19.62 & 0.8200 & 0.1455 & 16.21 & 0.7197 & 0.2181 & 16.40 & 0.6189 & 0.2422 \\
 &~NeRF-W\tnote{*} & 24.00 & 0.8758 & 0.1332 & 21.07 & \textbf{0.8422} & \secondbest{0.1119} & 19.75 & 0.7207 &  0.2029\\
 &~Ha-NeRF & 24.58 & 0.8829 & 0.0927 & 20.36 & 0.7947 & 0.1317 & \secondbest{20.27} & \secondbest{0.7270} & \secondbest{0.1628}  \\
 & \cellcolor[rgb]{ .949,  .863,  .859}~\sexyname-R~(Ours)~ & \cellcolor[rgb]{ .949,  .863,  .859}\secondbest{26.18} & \cellcolor[rgb]{ .949,  .863,  .859}\secondbest{0.8937} & \cellcolor[rgb]{ .949,  .863,  .859}\secondbest{0.0840} & \cellcolor[rgb]{ .949,  .863,  .859}\secondbest{21.64} &	\cellcolor[rgb]{ .949,  .863,  .859}0.8206 &	\cellcolor[rgb]{ .949,  .863,  .859}0.1160      & \cellcolor[rgb]{ .949,  .863,  .859}20.08 & \cellcolor[rgb]{ .949,  .863,  .859}0.6538 & \cellcolor[rgb]{ .949,  .863,  .859}0.2372 \\
  & \cellcolor[rgb]{ .949,  .863,  .859}~\sexyname~(Ours)~& \cellcolor[rgb]{ .949,  .863,  .859}\textbf{26.86} & \cellcolor[rgb]{ .949,  .863,  .859}\textbf{0.9069} & \cellcolor[rgb]{ .949,  .863,  .859}\textbf{0.0733} & \cellcolor[rgb]{ .949,  .863,  .859}\textbf{22.03} & \cellcolor[rgb]{ .949,  .863,  .859}\secondbest{0.8369} & \cellcolor[rgb]{ .949,  .863,  .859}\textbf{0.1060} &\cellcolor[rgb]{ .949,  .863,  .859}\textbf{22.02} 	    & \cellcolor[rgb]{ .949,  .863,  .859}\textbf{0.7488} 	 &\cellcolor[rgb]{ .949,  .863,  .859}\textbf{0.1354}  \\
 \midrule
 \multicolumn{1}{c|}{\multirow{5}[2]{*}{\rotatebox{90}{R / 2}}}&~NeRF & 18.90 & 0.8159 & 0.2316 & 15.60 & 0.7155 & 0.2916 & 16.14 & 0.6007 & 0.3662 \\
 &~NeRF-W\tnote{*} & 24.17 & 0.8905 & 0.1670 & 19.20 & 0.8076 & 0.1915 & 18.97 & 0.6984 & 0.2652 \\
 &~Ha-NeRF & 24.04 & 0.8773 & 0.1391 & 20.02 & 0.8012 & 0.1710 & 20.18 & 0.6908 & 0.2225 \\
 &\cellcolor[rgb]{ .949,  .863,  .859}~\sexyname-R~(Ours)~ & \cellcolor[rgb]{ .949,  .863,  .859}\secondbest{25.94} & \cellcolor[rgb]{ .949,  .863,  .859}\secondbest{0.8929} & \cellcolor[rgb]{ .949,  .863,  .859}\secondbest{0.1378} & \cellcolor[rgb]{ .949,  .863,  .859}\secondbest{21.66} & \cellcolor[rgb]{ .949,  .863,  .859}\secondbest{0.8171} & \cellcolor[rgb]{ .949,  .863,  .859}\secondbest{0.1646} & \cellcolor[rgb]{ .949,  .863,  .859}\secondbest{21.37} &	\cellcolor[rgb]{ .949,  .863,  .859}\secondbest{0.7111} 	& \cellcolor[rgb]{ .949,  .863,  .859}\secondbest{0.2212}      \\
& \cellcolor[rgb]{ .949,  .863,  .859}~\sexyname~(Ours)~ & \cellcolor[rgb]{ .949,  .863,  .859}\textbf{26.53}  & \cellcolor[rgb]{ .949,  .863,  .859}\textbf{0.9003} & \cellcolor[rgb]{ .949,  .863,  .859}\textbf{0.1060} & \cellcolor[rgb]{ .949,  .863,  .859}\textbf{22.07} & \cellcolor[rgb]{ .949,  .863,  .859}\textbf{0.8233} & \cellcolor[rgb]{ .949,  .863,  .859}\textbf{0.1520} & \cellcolor[rgb]{ .949,  .863,  .859}\textbf{21.48} & \cellcolor[rgb]{ .949,  .863,  .859}\textbf{0.7117} & \cellcolor[rgb]{ .949,  .863,  .859}\textbf{0.2069}\\
 \bottomrule
\end{tabular}
}
	 \end{threeparttable}
	 \end{center}
\caption{Quantitative experimental results on three real-world datasets under two resolution settings, \ie~downscaling original image resolution by 2 (R/2) and 4 (R/4). The \textbf{bold} and the \secondbest{underlined} numbers indicate the best and second-best results, respectively.} 
\label{tab:comparison}
\end{table*}

To deal with transient objects caused by unconstrained photo collections for novel-view synthesis, we propose a new perspective, \ie~obtaining the visibility map of transient objects by segmenting the reference image $\mathcal{I}_{a}$. With the receptive fields of a deep segmentation network~\cite{luo2016understanding}, the interactions of different pixels and rays are facilitated, thus introducing more global information. 
\par
\begin{figure}
\centering
\includegraphics[width=0.45\textwidth]{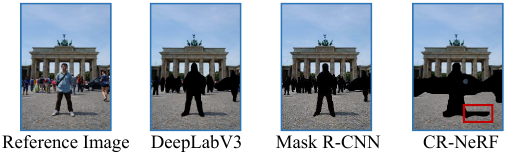}
\captionof{figure}{Transient objects from pre-trained DeeplabV3~\cite{deeplabv32018}, pre-trained Mask R-CNN~\cite{he2017mask} and \sexyname. \sexyname~captures a shadow of a tourist without using segmentation labels for training.}
\label{fig:mask}
\end{figure}

 To accurately detect transient objects,  we start by exploring a pre-trained Mask R-CNN model~\cite{he2017mask} and a pre-trained DeepLabV3 model~\cite{deeplabv32018} that are capable of effectively segmenting common objects such as tourists and cars, \etc~We observe that although the models properly segment the common objects, the reconstruction error is amplified (see empirical studies in Sect.~\ref{sec:ablation}). 
The possible reason is that the target transient objects are not limited to common objects, more objects 
 (\eg~shadows of tourists in Fig.~\ref{fig:mask}) should also be taken into consideration.
 \par In this sense, we choose a learning-based manner to select which objects to segment and therefore deploy a light-weight segmentation network $\mathcal{S}_{\theta_{\Delta}}$ following~\cite{wu2020cgnet}. Since we cannot sample all rays that interact with $\mathcal{I}_{a}$ due to limited GPU memory in the training phase, naively processing all rays of transient objects (\ie~$\mS_{\theta_{\Delta}}(\mathcal{I}_{a})$) is therefore not applicable. Hence, we apply a grid sampling strategy (GS)~\cite{Schwarz2020NEURIPS}  which samples $\mS_{\theta_{\Delta}}(\mathcal{I}_{a})$ to pair with $m$ rays (see Fig.~\ref{fig:overview}).  The whole process for estimating $\textbf{M}$ is:
\begin{equation}
\label{eqn:generate_mask}
\begin{aligned}
    \textbf{M} = \mathrm{GS}(\mS_{\theta_{\Delta}}(\mathcal{I}_{a})),
   \end{aligned}
\end{equation} 
where $\mS_{\theta_{\Delta}}: \mathbb{R}^{3 \times h_{cr1}h_{cr2}  }  \to \mathbb{R}^{3 \times h_{cr1}h_{cr2}}$, $h_{cr1}$ and $h_{cr2}$ are heights and width of $\mathcal{I}_{a}$. Here, $\mathcal{S}_{\theta_{\Delta}}$ learns a visibility map $\textbf{M}$ without the supervision of ground truth segmentation masks. During training, we set $m$ to be smaller than $h_{cr1}h_{cr2}$ for saving computational overhead.

\textbf{Loss function $\mL_t$ for eliminating transient objects:} The loss function for handling transient objects is:
\begin{equation}
\label{loss: transient}
    \mathcal{L}_{t} = ||(1-\textbf{M}) \odot (\mathcal{I}_{\mathrm{n}}-\mathcal{I}_{a})^{2}||_1+\lambda_{0}\|{\mathbf{M}}\|^{2}, 
\end{equation}
where $\odot$ denotes element-wise multiplication. The loss $\mathcal{L}_{t}$ aims to mask out transient objects via $\textbf{M}$. To prevent our transient network from masking everything, we follow Ha-NeRF to add $\lambda_{0}\|{\mathbf{M}}\|^{2}$ as a regularization term. 

\subsection{Difference of \sexyname~with Existing Methods}   
\par To model varying appearances, Ha-NeRF and NeRF-W process each single ray independently by Eqn.~\ref{eq:style_previous}. To handle transient objects, NeRF-W implements an additional MLP for rendering transient objects by Eqn.~\ref{eq:style_previous}. Ha-NeRF estimates a visibility map by separately utilizing a UV coordinate and a conditional feature of a reference image.   Differently,~\sexyname~considers information across multiple rays. Specifically,~\sexyname~takes $m$ rays as input, fuses them with a conditional feature, and generates a region of an image simultaneously (see Fig.~\ref{fig:movivation} (b)). A recent work \ie,~4K-NeRF To capture ray correlation, leverages depth-modulated convolutions. In contrast, \sexyname~captures the covariance of different rays. We theoretically (\cf~Sec.~\ref{sec:appearance}) and empirically (see details in the appendix) analyze the necessity of considering multiple rays.

\begin{figure*}[th]
    \centering
\includegraphics[width=1\textwidth]{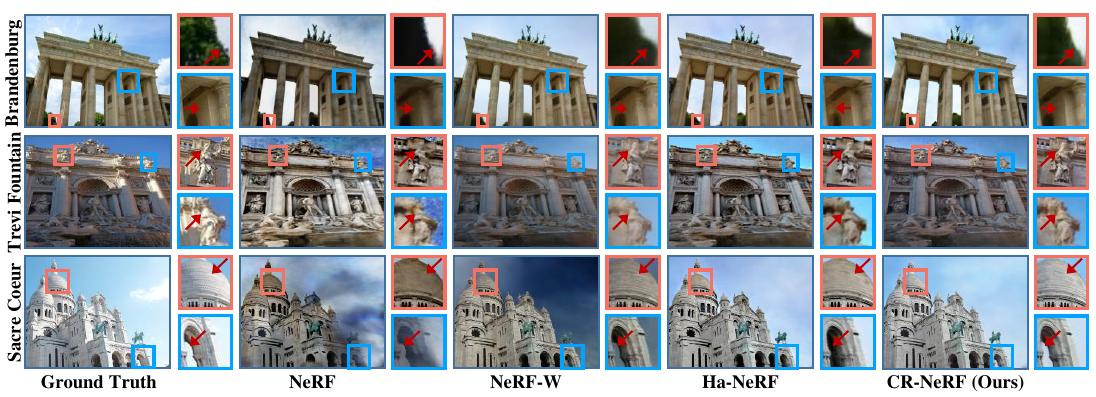}

    \caption{Qualitative experimental results on three unconstrained datasets. \sexyname~recovers realistic appearance (\eg~green plant in Brandenburg, sunshine on statues in Trevi, and light blue sky in Sacre.). Moreover, \sexyname~removes transient objects for a consistent geometry (\eg~less ghost effects around pillars of Brandenburg and Sacre).}
    \label{fig:comparison_all}
\end{figure*}

\begin{figure*}[th]
    \centering
\includegraphics[width=1\textwidth]{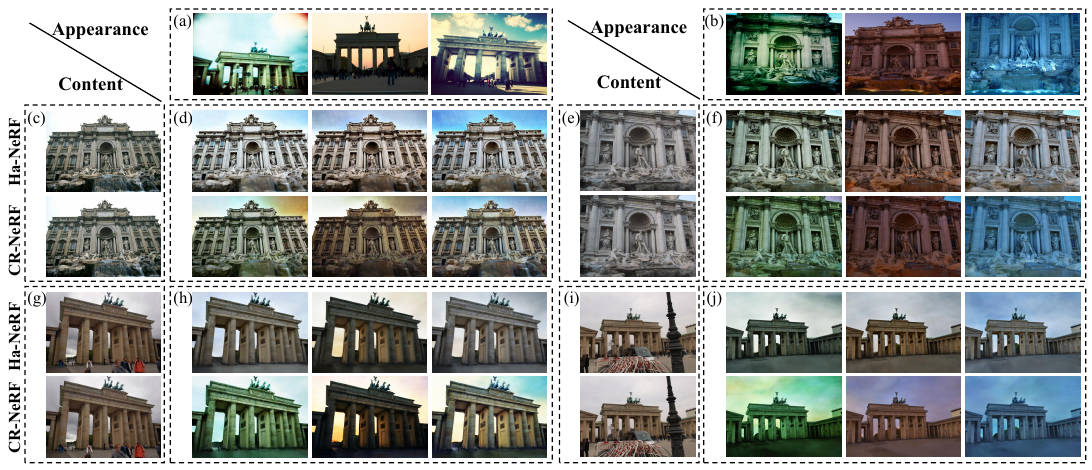}
    \caption{Modeling appearance in Brandenburg and Trevi datasets using various viewing directions and appearance images. The viewing directions of the synthesized images are the same as that of the nearest content images on the left. (a, b) appearance images from Brandenburg and Trevi, respectively. (c, e) Content images from Trevi. (g, i) Content image from Brandenburg. (d, f) Synthesized images in Trevi. (h, j) Synthesized images in Brandenburg. }
    \label{fig:transfer_style}
\end{figure*}

\section{Experiments}
\par \textbf{Implementation details.} 
We implement our approach using PyTorch~\cite{paszke2019pytorch} and train our networks with Adam~\cite{kingma2014adam} optimizer. For a fair comparison, we follow all common hyper-parameter settings same as Ha-NeRF~\cite{chen2022hallucinated}, \eg~setting the number of input rays, learning rate, $\lambda$ and height and width of fully connected layers to $1024$, $5\times10^{-4}$, $1\times10^{-3}$, $8$ and $256$, respectively. We set $\beta$ to $1\times10^{-5}$.
For a thorough study, we downscale the original image of each dataset by 2 times (R/2) and 4 times (R/4). 
During inference, we omit the segmentation network $\mathcal{S}_{\theta_{\Delta}}$, see more details about the inference of \sexyname~in the appendix. The code and data can be found at \url{https://github.com/YifYang993/CR-NeRF-PyTorch.git}.
\par \textbf{Datasets, metrics, and comparison methods.} 
Following Ha-NeRF~\cite{chen2022hallucinated}, we evaluate our proposed method on three datasets: Brandenburg Gate, Sacre Coeur, and Trevi Fountain. For visual inspection, we present rendered images generated from the same set of input views. We also report quantitative results based on PSNR, SSIM~\cite{wang2004image}, and LPIPS~\cite{zhang2018unreasonable, huang2022agtgan}. We evaluate our proposed method against NeRF~\cite{mildenhall2021nerf}, NeRF-W~\cite{martinbrualla2020nerfw}, Ha-NeRF~\cite{chen2022hallucinated}.
For ablation studies, we construct several variants of our CR-NeRF: 
1) \sexyname-R replaces the cross-ray features from \sexyname~with a ray-point-level features, \ie~features of ray points along multiple rays;
2) \sexyname-B is constructed upon \sexyname~without the cross-rays appearance modeling module and transient handling module;
3) \sexyname-A eliminates the cross-rays appearance modeling module only and
4) \sexyname-T removes the transient handling module.

\subsection{Comparison Experiments} 
\par \textbf{Quantitative experiments.} We conduct extensive experiments on Brandenburg Gate, Sacre Coeur, and Trevi Fountain datasets. We follow Ha-NeRF with the image resolution setting of 2$\times$ downscaling~(R/2) and further evaluate the effectiveness of our \sexyname~on  4$\times$ downscaling~(R/4). As demonstrated in Tab.~\ref{tab:comparison}, we observe that vanilla NeRF performs worst among all methods, since NeRF assumes the scene behind the training images is static. By modeling the style embedding and handling the transient objects, NeRF-W and Ha-NeRF achieve competitive performance in terms of PSNR, SSIM, and LPIPS. Note that NeRF-W optimizes its style embedding on test images since NeRF-W can not transfer to unseen test images directly. Thus, the comparison with NeRF-W is unfair. Even with the unfair comparison, thanks to the cross-ray manner, our \sexyname~outperforms NeRF-W and Ha-NeRF on Brandenburg and Trevi under two downscaling settings.  
\par \textbf{Qualitative experiments.} We summarize the qualitative results of all comparison methods in Fig.~\ref{fig:comparison_all}. We observe that NeRF produces foggy artifacts and inaccurate appearance. NeRF-W and Ha-NeRF are able to reconstruct a more promising 3D geometry and model appearance from the ground truth image. However, the reconstructed geometry is not accurate enough, \eg~the shape of the green plant and ghost effects around the pillar in Brandenburg, the cavity in Sacre. Besides, the transferred appearance is not realistic enough, \eg~sunshine on statues in Sacre, and the color of blue sky and grey roof in Trevi. Differently, our \sexyname~introduces a cross-ray paradigm and therefore achieves more realistic appearance modeling and reconstructs a consistent geometry by suppressing transient objects.    
\par \textbf{Comparison of appearance modeling.} We investigate the appearance modeling ability of our \sexyname~in Fig.~\ref{fig:transfer_style}. We observe that 1) \sexyname~captures appearance information more accurately than Ha-NeRF, especially towards recovering appearances from images with high-frequency information, \eg~green sky, blue sky, red building, sunlight on the gate. 2) \sexyname~successfully removes transient objects such as tourists and cars while retaining static objects such as roads and buildings.      

\begin{table*}[t]
\centering
\small
 \begin{threeparttable}
\setlength{\tabcolsep}{1.5pt} 
\begin{tabular}{l|ccccccccc}
\toprule
\multicolumn{1}{l}{} & \multicolumn{3}{c}{Brandenburg Gate} & \multicolumn{3}{c}{Sacre Coeur} & \multicolumn{3}{c}{Trevi Fountain} \\
\cmidrule(lr){2-4} \cmidrule(lr){5-7} \cmidrule(lr){8-10}
\multicolumn{1}{l}{} &~~PSNR ($\uparrow$) &~~SSIM ($\uparrow$) &~~LPIPS ($\downarrow$) &~~PSNR ($\uparrow$) &~~SSIM ($\uparrow$) &~~LPIPS ($\downarrow$) &~~PSNR ($\uparrow$) &~~SSIM ($\uparrow$) &~~LPIPS ($\downarrow$) \\    
\midrule
\sexyname-B~           & 19.58      & 0.8216     & 0.1470      & 16.11    & 0.7145    & 0.2196    & 16.37     & 0.6206     & 0.2493    \\
\sexyname-A~            & 26.38      & 0.8929     & 0.0885     & 21.67    & 0.8182    & 0.1127    & 21.85     & 0.7473     & 0.1388    \\
\sexyname-T~            & 20.46      & 0.8361     & 0.1300     & 16.28    & 0.7650    & 0.1799    & 16.55     & 0.6446     & 0.2230    \\
\sexyname~              & \textbf{26.86}      & \textbf{0.9069}     & \textbf{0.0733}     & \textbf{22.03}    & \textbf{0.8369}    & \textbf{0.1060}    & \textbf{22.02}     & \textbf{0.7488}     & \textbf{0.1354}   \\
\bottomrule
\end{tabular}
\end{threeparttable}
\caption{Ablation studies of \sexyname~on three datasets. The performance of our baseline (\sexyname-B) is progressively improved by adding the appearance modeling module (\sexyname-A) and the transient handler (\sexyname-T). The bold numbers indicate the best result.} 
\label{tab:ablation}
\end{table*}

\subsection{Ablation Studies}
\label{sec:ablation}

\par \textbf{Ablation of appearance module and transient module.}
We summarize the ablation studies of \sexyname~on Brandenburg, Sacre, and Trevi dataset in Tab.~\ref{tab:ablation}. We observe that \sexyname-A and \sexyname-T outperform \sexyname-B. and \sexyname~exceeds all variants, indicating the effectiveness of our Appearance Module and Transient Module.  

\par \textbf{Cross-ray manner and fusing level.} We study the effectiveness of the cross-ray manner and the fusing level by comparing with our baseline \sexyname-R quantitatively in Tab.~\ref{tab:comparison} and qualitatively in the appendix. From Tab.~\ref{tab:comparison}, \sexyname-R achieves a competitive performance on three datasets, which shows the superiority of leveraging various rays. Moreover, our proposed \sexyname~outperforms \sexyname-R consistently on all datasets. We assume that compared with the cross-ray-point features, the granularity of the cross-ray features $\mathcal{F}^{\mathrm{cr}}$ is closer to that of the image-level conditional features. Therefore, feature fusion is more effective. we provide qualitative results in the appendix.
\subsection{Further Experiments}
\par
\begin{figure}[t]
    \centering
\includegraphics[width=0.48\textwidth]{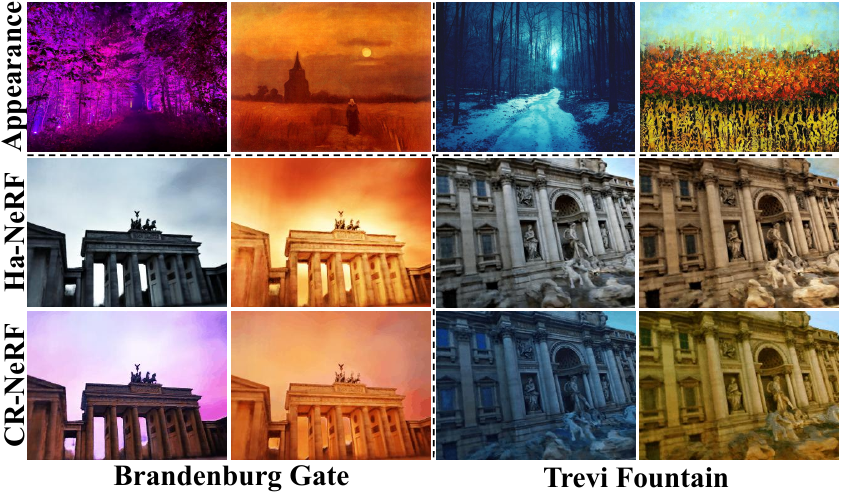}
    \caption{Modeling appearance from unseen images with high-frequency information to Brandenburg and Trevi.}
    \label{fig:unseen_style_small}
\end{figure}

 \textbf{Unseen appearance modeling.} 
Our proposed \sexyname~is able to deal with unseen appearance images thanks to the ability of our cross-ray appearance modeling handler.  As shown in Fig.~\ref{fig:unseen_style_small}, our \sexyname~captures the whole range appearance (\eg~the blue and purple appearance in the last two columns in Brandenburg and Trevi fountain datasets) of the given style image more accurately compared with Ha-NeRF. Moreover,  our \sexyname~synthesizes a more consistent appearance than images generated by Ha-NeRF (\eg~the sudden bright light on the sky of the second column in Brandenburg dataset). Note that NeRF-W needs to optimize its appearance embedding on each test image by pixel-level supervision, thus NeRF-W cannot be directly applied to unseen appearance modeling.

\par \textbf{Inference time on multiple images.} When dealing with multiple images of various appearances with fixed camera position, the inference efficiency of our \sexyname~exceeds Ha-NeRF significantly (\ie~$2.12$ seconds vs $24.09$ seconds in Tab.~\ref{tab:inferencetime}). The reason is that our \sexyname~generates cross-ray features $\mathcal{F}^{\mathrm{cr}}$ only once by using a NeRF backbone and synthesizes various appearances by fusing $\mathcal{F}^{\mathrm{cr}}$ and appearance embedding of each image. In contrast, Ha-NeRF requires the use of its NeRF backbone for each estimation. For efficiency, we modify Ha-NeRF by saving its interim results. However, since the interim results of Ha-NeRF occupy a large amount of GPU memory beyond the capacity of the single TITAN Xp GPU, moving the results to the host memory requires additional I/O time.

\begin{table}[t]
\centering
\resizebox{1\linewidth}{!}{
\begin{tabular}{ccccc}
\toprule
\multicolumn{1}{c}{(Seconds per Image)} & \multicolumn{2}{c}{A Single Image} & \multicolumn{2}{c}{Multiple Images} \\
\cmidrule(lr){2-3} \cmidrule(lr){4-5} 
 & R/4 & R/2 & R/4 & R/2 \\
 \midrule
Ha-NeRF & 4.92s & 14.01s & 3.88s & 24.09s \\
\sexyname~(Ours) & 4.02s & 15.07s & 0.92s & 2.12s\\
\bottomrule
\end{tabular}
}
\caption{Inference time of \sexyname~and Ha-NeRF with one TITAN Xp GPU on Brandenburg with two downscaling ratios.}
\label{tab:inferencetime}%
\end{table}

\par \textbf{Transient objects handling.} We observe that simply masking common objects harms reconstruction performance. Specifically, we use a pre-trained DeepLabV3 and a pre-trained Mask R-CNN that produce promising segmentation results for common objects such as pedestrians and cars~(we carefully choose the categories for estimation to avoid masking out static objects). However, performance degrades when combining \sexyname-A with these two networks (see Tab.~\ref{tab:mask}). Considering that the transient handler of \sexyname~is trained without the supervision of ground truth visibility maps, our estimated visibility maps are inevitably less accurate than the pre-trained network on the common objects (see the appendix for more details). We assume the definition of transient objects is still an open question and we leave it to our future work. 

\begin{table}[t]
\centering
 \begin{threeparttable}
 \LARGE
\resizebox{0.95\linewidth}{!}{
\begin{tabular}{l|ccc}
\toprule
            & \multicolumn{1}{l}{~~PSNR ($\uparrow$)} & \multicolumn{1}{l}{~~SSIM ($\uparrow$)} & \multicolumn{1}{l}{~~LPIPS ($\downarrow$)} \\
            \midrule
            \sexyname-A            & 26.38      & 0.8929     & 0.0885 \\
            \midrule
\sexyname-A + DeepLabV3 & $\text{24.89}_{({\downarrow}1.49)}$                    & $\text{0.8781}_{({\downarrow}0.0148)}$                  & $\text{0.1065}_{({\uparrow}0.0180)}$                    \\
 \sexyname-A + Mask R-CNN  & $\text{25.46}_{({\downarrow} 0.92)} $                   & $\text{0.8885}_{({\downarrow} 0.0044)}$                   & $\text{0.0919}_{({\uparrow} 0.0034)}$                     \\ 
\sexyname & $\textbf{26.86}_{({\uparrow} 0.48)}$      & $\textbf{0.9069}_{({\uparrow} 0.0140)}$     & $\textbf{0.0733}_{({\downarrow} 0.0152)}$  \\    
\bottomrule
\end{tabular}}
\end{threeparttable}
\caption{Discussion on transient handlers on Brandenburg dataset.}
\label{tab:mask}
\end{table}

\section{Conclusion}
In this paper, we address novel-view synthesis from unconstrained images by considering the information of multiple rays within a scene. The unconstrained scenario introduces the varying appearances and transient objects in the images. We propose a novel cross-ray paradigm for the task by leveraging global interactive information across multiple rays. Guided by the paradigm, to address the variable appearance, we propose to represent information of multiple rays with cross-ray features and then inject an appearance of each image via fuse feature covariance of the rays and the image appearance. To handle transient objects, we propose a novel perspective of handling transient objects via image segmentation on multiple rays. Based on this, we estimate and grid sample a visibility map to pair with the rays. Extensive experimental results on large real-world datasets show the effectiveness of our proposed method.  

\section*{Acknowledgements}
This work was partially supported by the 
National Natural Science Foundation of China (NSFC) (62072190), National Natural Science Foundation of China (NSFC) 61836003 (key project), Program for Guangdong Introducing Innovative and Enterpreneurial Teams 2017ZT07X183.

{\small
\bibliographystyle{ieee_fullname}
\bibliography{egbib}

\begin{thebibliography}{10}\itemsep=-1pt

\bibitem{barron2022mip}
Jonathan~T Barron, Ben Mildenhall, Dor Verbin, Pratul~P Srinivasan, and Peter
  Hedman.
\newblock Mip-nerf 360: Unbounded anti-aliased neural radiance fields.
\newblock In {\em CVPR}, pages 5470--5479, 2022.

\bibitem{buehler2001unstructured}
Chris Buehler, Michael Bosse, Leonard McMillan, Steven Gortler, and Michael
  Cohen.
\newblock Unstructured lumigraph rendering.
\newblock In {\em SIGGRAPH}, pages 425--432, 2001.

\bibitem{chen2021mvsnerf}
Anpei Chen, Zexiang Xu, Fuqiang Zhao, Xiaoshuai Zhang, Fanbo Xiang, Jingyi Yu,
  and Hao Su.
\newblock Mvsnerf: Fast generalizable radiance field reconstruction from
  multi-view stereo.
\newblock In {\em ICCV}, pages 14124--14133, 2021.

\bibitem{deeplabv32018}
Liang-Chieh Chen, George Papandreou, Florian Schroff, and Hartwig Adam.
\newblock Rethinking atrous convolution for semantic image segmentation.
\newblock In {\em CVPR}, page 4067–4075, 2017.

\bibitem{chen1993view}
Shenchang~Eric Chen and Lance Williams.
\newblock View interpolation for image synthesis.
\newblock In {\em SIGGRAPH}, pages 279--288, 1993.

\bibitem{chen2022hallucinated}
Xingyu Chen, Qi Zhang, Xiaoyu Li, Yue Chen, Ying Feng, Xuan Wang, and Jue Wang.
\newblock Hallucinated neural radiance fields in the wild.
\newblock In {\em CVPR}, pages 12943--12952, 2022.

\bibitem{cheng2008improved}
Chia-Ming Cheng, Shu-Jyuan Lin, Shang-Hong Lai, and Jinn-Cherng Yang.
\newblock Improved novel view synthesis from depth image with large baseline.
\newblock In {\em ICPR}, pages 1--4, 2008.

\bibitem{dai2023disentangling}
Gang Dai, Yifan Zhang, Qingfeng Wang, Qing Du, Zhuliang Yu, Zhuoman Liu, and
  Shuangping Huang.
\newblock Disentangling writer and character styles for handwriting generation.
\newblock In {\em Proceedings of the IEEE/CVF Conference on Computer Vision and
  Pattern Recognition}, pages 5977--5986, 2023.

\bibitem{drebin1988volume}
Robert~A Drebin, Loren Carpenter, and Pat Hanrahan.
\newblock Volume rendering.
\newblock In {\em SIGGRAPH}, pages 65--74, 1988.

\bibitem{fan2017point}
Haoqiang Fan, Hao Su, and Leonidas~J Guibas.
\newblock A point set generation network for 3d object reconstruction from a
  single image.
\newblock In {\em CVPR}, pages 605--613, 2017.

\bibitem{flynn2019deepview}
John Flynn, Michael Broxton, Paul Debevec, Matthew DuVall, Graham Fyffe, Ryan
  Overbeck, Noah Snavely, and Richard Tucker.
\newblock Deepview: View synthesis with learned gradient descent.
\newblock In {\em CVPR}, pages 2367--2376, 2019.

\bibitem{gatys2016image}
Leon~A Gatys, Alexander~S Ecker, and Matthias Bethge.
\newblock Image style transfer using convolutional neural networks.
\newblock In {\em CVPR}, pages 2414--2423, 2016.

\bibitem{guo2021ad}
Yudong Guo, Keyu Chen, Sen Liang, Yong-Jin Liu, Hujun Bao, and Juyong Zhang.
\newblock Ad-nerf: Audio driven neural radiance fields for talking head
  synthesis.
\newblock In {\em CVPR}, pages 5784--5794, 2021.

\bibitem{he2017mask}
Kaiming He, Georgia Gkioxari, Piotr Doll{\'a}r, and Ross Girshick.
\newblock Mask r-cnn.
\newblock In {\em ICCV}, pages 2961--2969, 2017.

\bibitem{henzler2019escaping}
Philipp Henzler, Niloy~J Mitra, and Tobias Ritschel.
\newblock Escaping plato's cave: 3d shape from adversarial rendering.
\newblock In {\em ICCV}, pages 9984--9993, 2019.

\bibitem{huang2022agtgan}
Hongxiang Huang, Daihui Yang, Gang Dai, Zhen Han, Yuyi Wang, Kin-Man Lam, Fan
  Yang, Shuangping Huang, Yongge Liu, and Mengchao He.
\newblock Agtgan: Unpaired image translation for photographic ancient character
  generation.
\newblock In {\em Proceedings of the 30th ACM international conference on
  multimedia}, pages 5456--5467, 2022.

\bibitem{jiang2018gal}
Li Jiang, Shaoshuai Shi, Xiaojuan Qi, and Jiaya Jia.
\newblock Gal: Geometric adversarial loss for single-view 3d-object
  reconstruction.
\newblock In {\em ECCV}, pages 802--816, 2018.

\bibitem{jiang2023alignerf}
Yifan Jiang, Peter Hedman, Ben Mildenhall, Dejia Xu, Jonathan~T Barron,
  Zhangyang Wang, and Tianfan Xue.
\newblock Alignerf: High-fidelity neural radiance fields via alignment-aware
  training.
\newblock In {\em Proceedings of the IEEE/CVF Conference on Computer Vision and
  Pattern Recognition}, pages 46--55, 2023.

\bibitem{jiang2020sdfdiff}
Yue Jiang, Dantong Ji, Zhizhong Han, and Matthias Zwicker.
\newblock Sdfdiff: Differentiable rendering of signed distance fields for 3d
  shape optimization.
\newblock In {\em CVPR}, pages 1251--1261, 2020.

\bibitem{kingma2014adam}
Diederik~P Kingma and Jimmy Ba.
\newblock Adam: A method for stochastic optimization.
\newblock In {\em ICLR}, 2015.

\bibitem{levoy1996light}
Marc Levoy and Pat Hanrahan.
\newblock Light field rendering.
\newblock In {\em SIGGRAPH}, pages 31--42, 1996.

\bibitem{li2022neural}
Tianye Li, Mira Slavcheva, Michael Zollhoefer, Simon Green, Christoph Lassner,
  Changil Kim, Tanner Schmidt, Steven Lovegrove, Michael Goesele, Richard
  Newcombe, et~al.
\newblock Neural 3d video synthesis from multi-view video.
\newblock In {\em Proceedings of the IEEE/CVF Conference on Computer Vision and
  Pattern Recognition}, pages 5521--5531, 2022.

\bibitem{li2019learning}
Xueting Li, Sifei Liu, Jan Kautz, and Ming-Hsuan Yang.
\newblock Learning linear transformations for fast image and video style
  transfer.
\newblock In {\em CVPR}, pages 3809--3817, 2019.

\bibitem{li2017universal}
Yijun Li, Chen Fang, Jimei Yang, Zhaowen Wang, Xin Lu, and Ming{-}Hsuan Yang.
\newblock Universal style transfer via feature transforms.
\newblock In {\em NeurIPS}, pages 386--396, 2017.

\bibitem{li2021neural}
Zhengqi Li, Simon Niklaus, Noah Snavely, and Oliver Wang.
\newblock Neural scene flow fields for space-time view synthesis of dynamic
  scenes.
\newblock In {\em Proceedings of the IEEE/CVF Conference on Computer Vision and
  Pattern Recognition}, pages 6498--6508, 2021.

\bibitem{lin2018learning}
Chen-Hsuan Lin, Chen Kong, and Simon Lucey.
\newblock Learning efficient point cloud generation for dense 3d object
  reconstruction.
\newblock In {\em AAAI}, pages 7114--7121, 2018.

\bibitem{liu2020dist}
Shaohui Liu, Yinda Zhang, Songyou Peng, Boxin Shi, Marc Pollefeys, and Zhaopeng
  Cui.
\newblock Dist: Rendering deep implicit signed distance function with
  differentiable sphere tracing.
\newblock In {\em CVPR}, pages 2019--2028, 2020.

\bibitem{lo2018food}
Frank P-W Lo, Yingnan Sun, Jianing Qiu, and Benny Lo.
\newblock Food volume estimation based on deep learning view synthesis from a
  single depth map.
\newblock {\em Nutrients}, 10(12):2005, 2018.

\bibitem{lu2019closed}
Ming Lu, Hao Zhao, Anbang Yao, Yurong Chen, Feng Xu, and Li Zhang.
\newblock A closed-form solution to universal style transfer.
\newblock In {\em ICCV}, pages 5952--5961, 2019.

\bibitem{luo2016understanding}
Wenjie Luo, Yujia Li, Raquel Urtasun, and Richard Zemel.
\newblock Understanding the effective receptive field in deep convolutional
  neural networks.
\newblock In {\em NeurIPS}, 2016.

\bibitem{martinbrualla2020nerfw}
Ricardo Martin-Brualla, Noha Radwan, Mehdi S.~M. Sajjadi, Jonathan~T. Barron,
  Alexey Dosovitskiy, and Daniel Duckworth.
\newblock {NeRF in the Wild: Neural Radiance Fields for Unconstrained Photo
  Collections}.
\newblock In {\em CVPR}, pages 7210--7219, 2021.

\bibitem{mescheder2019occupancy}
Lars Mescheder, Michael Oechsle, Michael Niemeyer, Sebastian Nowozin, and
  Andreas Geiger.
\newblock Occupancy networks: Learning 3d reconstruction in function space.
\newblock In {\em CVPR}, pages 4460--4470, 2019.

\bibitem{mildenhall2021nerf}
Ben Mildenhall, Pratul~P Srinivasan, Matthew Tancik, Jonathan~T Barron, Ravi
  Ramamoorthi, and Ren Ng.
\newblock Nerf: Representing scenes as neural radiance fields for view
  synthesis.
\newblock In {\em ECCV}, pages 405--421, 2021.

\bibitem{olkin1982distance}
Ingram Olkin and Friedrich Pukelsheim.
\newblock The distance between two random vectors with given dispersion
  matrices.
\newblock {\em Linear Algebra Appl.}, 48:257--263, 1982.

\bibitem{park2019deepsdf}
Jeong~Joon Park, Peter Florence, Julian Straub, Richard Newcombe, and Steven
  Lovegrove.
\newblock Deepsdf: Learning continuous signed distance functions for shape
  representation.
\newblock In {\em CVPR}, pages 165--174, 2019.

\bibitem{park2021nerfies}
Keunhong Park, Utkarsh Sinha, Jonathan~T Barron, Sofien Bouaziz, Dan~B Goldman,
  Steven~M Seitz, and Ricardo Martin-Brualla.
\newblock Nerfies: Deformable neural radiance fields.
\newblock In {\em ICCV}, pages 5865--5874, 2021.

\bibitem{paszke2019pytorch}
Adam Paszke, Sam Gross, Francisco Massa, Adam Lerer, James Bradbury, Gregory
  Chanan, Trevor Killeen, Zeming Lin, Natalia Gimelshein, Luca Antiga, et~al.
\newblock Pytorch: An imperative style, high-performance deep learning library.
\newblock In {\em NeurIPS}, pages 8024--8035, 2019.

\bibitem{peng2022tmvnet}
Kebin Peng, Rifatul Islam, John Quarles, and Kevin Desai.
\newblock Tmvnet: Using transformers for multi-view voxel-based 3d
  reconstruction.
\newblock In {\em CVPR}, pages 222--230, 2022.

\bibitem{Peng2020ECCV}
Songyou Peng, Michael Niemeyer, Lars Mescheder, Marc Pollefeys, and Andreas
  Geiger.
\newblock Convolutional occupancy networks.
\newblock In {\em ECCV}, pages 523--540, 2020.

\bibitem{pumarola2021d}
Albert Pumarola, Enric Corona, Gerard Pons-Moll, and Francesc Moreno-Noguer.
\newblock D-nerf: Neural radiance fields for dynamic scenes.
\newblock In {\em Proceedings of the IEEE/CVF Conference on Computer Vision and
  Pattern Recognition}, pages 10318--10327, 2021.

\bibitem{rockwell2021pixelsynth}
Chris Rockwell, David~F Fouhey, and Justin Johnson.
\newblock Pixelsynth: Generating a 3d-consistent experience from a single
  image.
\newblock In {\em ICCV}, pages 14104--14113, 2021.

\bibitem{schonberger2016structure}
Johannes~L Schonberger and Jan-Michael Frahm.
\newblock Structure-from-motion revisited.
\newblock In {\em CVPR}, pages 4104--4113, 2016.

\bibitem{Schwarz2020NEURIPS}
Katja Schwarz, Yiyi Liao, Michael Niemeyer, and Andreas Geiger.
\newblock Graf: Generative radiance fields for 3d-aware image synthesis.
\newblock In {\em NeurIPS}, pages 20154--20166, 2020.

\bibitem{srinivasan2021nerv}
Pratul~P Srinivasan, Boyang Deng, Xiuming Zhang, Matthew Tancik, Ben
  Mildenhall, and Jonathan~T Barron.
\newblock Nerv: Neural reflectance and visibility fields for relighting and
  view synthesis.
\newblock In {\em Proceedings of the IEEE/CVF Conference on Computer Vision and
  Pattern Recognition}, pages 7495--7504, 2021.

\bibitem{srinivasan2017learning}
Pratul~P Srinivasan, Tongzhou Wang, Ashwin Sreelal, Ravi Ramamoorthi, and Ren
  Ng.
\newblock Learning to synthesize a 4d rgbd light field from a single image.
\newblock In {\em ICCV}, pages 2243--2251, 2017.

\bibitem{su2021nerf}
Shih-Yang Su, Frank Yu, Michael Zollh{\"o}fer, and Helge Rhodin.
\newblock A-nerf: Articulated neural radiance fields for learning human shape,
  appearance, and pose.
\newblock In {\em NeurIPS}, pages 12278--12291, 2021.

\bibitem{tancik2022block}
Matthew Tancik, Vincent Casser, Xinchen Yan, Sabeek Pradhan, Ben Mildenhall,
  Pratul~P Srinivasan, Jonathan~T Barron, and Henrik Kretzschmar.
\newblock Block-nerf: Scalable large scene neural view synthesis.
\newblock In {\em Proceedings of the IEEE/CVF Conference on Computer Vision and
  Pattern Recognition}, pages 8248--8258, 2022.

\bibitem{turki2022mega}
Haithem Turki, Deva Ramanan, and Mahadev Satyanarayanan.
\newblock Mega-nerf: Scalable construction of large-scale nerfs for virtual
  fly-throughs.
\newblock In {\em Proceedings of the IEEE/CVF Conference on Computer Vision and
  Pattern Recognition}, pages 12922--12931, 2022.

\bibitem{turki2023suds}
Haithem Turki, Jason~Y Zhang, Francesco Ferroni, and Deva Ramanan.
\newblock Suds: Scalable urban dynamic scenes.
\newblock In {\em Proceedings of the IEEE/CVF Conference on Computer Vision and
  Pattern Recognition}, pages 12375--12385, 2023.

\bibitem{ullman1979interpretation}
Shimon Ullman.
\newblock The interpretation of structure from motion.
\newblock {\em Proceedings of the Royal Society of London. Series B. Biological
  Sciences}, 203(1153):405--426, 1979.

\bibitem{wang2021neus}
Peng Wang, Lingjie Liu, Yuan Liu, Christian Theobalt, Taku Komura, and Wenping
  Wang.
\newblock Neus: Learning neural implicit surfaces by volume rendering for
  multi-view reconstruction.
\newblock In {\em NeurIPS}, pages 27171--27183, 2021.

\bibitem{wang2004image}
Zhou Wang, Alan~C Bovik, Hamid~R Sheikh, and Eero~P Simoncelli.
\newblock Image quality assessment: from error visibility to structural
  similarity.
\newblock {\em IEEE TIP}, 13(4):600--612, 2004.

\bibitem{wang20224k}
Zhongshu Wang, Lingzhi Li, Zhen Shen, Li Shen, and Liefeng Bo.
\newblock 4k-nerf: High fidelity neural radiance fields at ultra high
  resolutions.
\newblock {\em arXiv preprint arXiv:2212.04701}, 2022.

\bibitem{Wilburn2005High}
Bennett Wilburn, Neel Joshi, Vaibhav Vaish, Eino-Ville Talvala, Emilio Antunez,
  Adam Barth, Andrew Adams, Mark Horowitz, and Marc Levoy.
\newblock High performance imaging using large camera arrays.
\newblock {\em ACM TOG}, 24(3):765–776, 2005.

\bibitem{wiles2020synsin}
Olivia Wiles, Georgia Gkioxari, Richard Szeliski, and Justin Johnson.
\newblock Synsin: End-to-end view synthesis from a single image.
\newblock In {\em CVPR}, pages 7467--7477, 2020.

\bibitem{wizadwongsa2021nex}
Suttisak Wizadwongsa, Pakkapon Phongthawee, Jiraphon Yenphraphai, and Supasorn
  Suwajanakorn.
\newblock Nex: Real-time view synthesis with neural basis expansion.
\newblock In {\em CVPR}, pages 8534--8543, 2021.

\bibitem{wu2020cgnet}
Tianyi Wu, Sheng Tang, Rui Zhang, Juan Cao, and Yongdong Zhang.
\newblock Cgnet: A light-weight context guided network for semantic
  segmentation.
\newblock {\em IEEE TIP}, 30:1169--1179, 2020.

\bibitem{xie2019pix2vox}
Haozhe Xie, Hongxun Yao, Xiaoshuai Sun, Shangchen Zhou, and Shengping Zhang.
\newblock Pix2vox: Context-aware 3d reconstruction from single and multi-view
  images.
\newblock In {\em ICCV}, pages 2690--2698, 2019.

\bibitem{xu2021generative}
Xudong Xu, Xingang Pan, Dahua Lin, and Bo Dai.
\newblock Generative occupancy fields for 3d surface-aware image synthesis.
\newblock In {\em NeurIPS}, pages 20683--20695, 2021.

\bibitem{zhang2021learning}
Jingyang Zhang, Yao Yao, and Long Quan.
\newblock Learning signed distance field for multi-view surface reconstruction.
\newblock In {\em ICCV}, pages 6525--6534, 2021.

\bibitem{zhang2018unreasonable}
Richard Zhang, Phillip Isola, Alexei~A Efros, Eli Shechtman, and Oliver Wang.
\newblock The unreasonable effectiveness of deep features as a perceptual
  metric.
\newblock In {\em CVPR}, pages 586--595, 2018.

\end{thebibliography}
}

\newpage

\onecolumn
\begin{table}[ht]
	\begin{tabular}{p{1\columnwidth}}
		\centering
		\Large{\textbf{Appendix for ``\mytitle''}}
	\end{tabular}
\end{table}

In the appendix, we provide detailed proofs of the proposition, more details, and more experimental results of the proposed Corss-Ray Neural Radiance Fields (\sexyname)\footnote{We suggest checking the video demo synthesized by our \sexyname~in the supplementary.}. We organize the appendix into the following sections. 
\begin{itemize}
    \item In Sec.~\ref{sec:sup_proof} we provides proof of our Proposition 1.
    \item In Sec.~\ref{sec:sup_Inference}, we provide details on inference of our \sexyname.
    \item In Sec.~\ref{sec:effectofrays}, we discuss the impact of the number of rays on our~\sexyname, which supports the necessity of considering multiple rays.
    \item In Sec.~\ref{sec:sup_Cross-Ray} we discuss the effectiveness of our cross-ray paradigm and fusing level.
    \item In Sec.~\ref{sec:sup_Interpolation} we demonstrate more synthesized views by interpolating between an appearance embedding to another.
    \item In Sec.~\ref{sec:sup_Appearance}, we report more qualitative experimental results of appearance modeling by comparing \sexyname~and existing methods on Brandenburg Gate and Trevi Fountain datasets.
    \item In Sec.~\ref{sec:sup_Unseen} we demonstrate more synthesized views by transferring appearance from unseen images.
    \item In Sec.~\ref{sec:training_time} we compare the training time of our \sexyname~with existing methods.
\end{itemize}

\setcounter{section}{0}
\renewcommand\thesection{\Alph{section}}
\makeatletter\@addtoreset{equation}{section}
\def\theequation{\thesection.\arabic{equation}}

\section{Proof of Proposition 1}
\label{sec:sup_proof}
\textbf{Proposition 1.}
\emph{
Given an invertible constant matrix $\bP {\in} \mathbb{R} ^{C {\times} C}$, assuming that $\mF^a {\sim} \mN(\boldsymbol{\mu}_a, \boldsymbol{\Sigma}_a)$,  $\mF^{\mathrm{cr}} {\sim} \mN(\boldsymbol{\mu}_{\mathrm{cr}}, \boldsymbol{\Sigma}_{\mathrm{cr}})$ and $\mT(\mF^{\mathrm{cr}}) {\sim} \mN(\boldsymbol{\mu}_a, \boldsymbol{\Sigma}_a)$, where $\mT(\mF^{\mathrm{cr}}) {=} \bT(\mF^{\mathrm{cr}} {-} \boldsymbol{\mu}_{\mathrm{cr}}) {+} \boldsymbol{\mu}_a$ and $\bT {\in} \mathbb{R}^ {C {\times} C}$ is a transformation matrix, the optimal $\bT$ to Problem (\ref{Prob: appearance modeling}) is:
\begin{equation}
\begin{aligned}
    \bT=\boldsymbol{\Sigma}_{\mathrm{cr}}^{-1 / 2}\left(\boldsymbol{\Sigma}_{\mathrm{cr}}^{1 / 2} \bP  \boldsymbol{\Sigma}_{a} \bP ^ \top\boldsymbol{\Sigma}_{\mathrm{cr}}^{1 / 2}\right)^{1 / 2} \boldsymbol{\Sigma}_{\mathrm{cr}}^{-1 / 2} \bP^{-1}.
\end{aligned}
\end{equation}
}
\begin{proof}
We rewrite Eqn. (\ref{Prob: appearance modeling}) 
using $\mT(\mF^{\mathrm{cr}}) {=} \bT(\mF^{\mathrm{cr}} {-} \boldsymbol{\mu}_{cs}) {+} \boldsymbol{\mu}_a$ as:
\begin{equation}
\label{eq: rewrite}
    \mmE_{\mF^{\mathrm{cr}}, \mF^a} \!\!\left[ \bT(\mF^{\mathrm{cr}} {-} \boldsymbol{\mu}_{\mathrm{cr}}) {+} \boldsymbol{\mu}_a {-}\mF^a \right]^ \top \!\left[ \bT(\mF^{\mathrm{cr}} {-} \boldsymbol{\mu}_{\mathrm{cr}}) {+} \boldsymbol{\mu}_a {-}\mF^a \right] {+} \beta \left\{\bP [\bT(\mF^{\mathrm{cr}} {-} \boldsymbol{\mu}_{\mathrm{cr}}) {+} \boldsymbol{\mu}_a] {-} \mF^{\mathrm{cr}}\right\}^\top \!\!\left\{\bP[ \bT(\mF^{\mathrm{cr}} {-} \boldsymbol{\mu}_{\mathrm{cr}}) {+} \boldsymbol{\mu}_a] {-} \mF^{\mathrm{cr}}\right\}.
\end{equation}
Let $\boldsymbol{u} = \mF^{\mathrm{cr}}{-} \boldsymbol{\mu}_{\mathrm{cr}}$, $\boldsymbol{v}=\bT\boldsymbol{u}$, $\boldsymbol{w} = \boldsymbol{\mu}_a {-}\mF^a$ and $\boldsymbol{\mu}_\Delta = \bP \boldsymbol{\mu}_a - \boldsymbol{\mu}_{\mathrm{cr}}$, based on $\mF^{\mathrm{cr}} {\sim} \mN(\boldsymbol{\mu}_{\mathrm{cr}}, \boldsymbol{\Sigma}_{\mathrm{cr}})$, $\mT(\mF^{\mathrm{cr}}) {\sim} \mN(\boldsymbol{\mu}_a, \boldsymbol{\Sigma}_a)$ and $\mF^a {\sim} \mN(\boldsymbol{\mu}_a, \boldsymbol{\Sigma}_a)$, we have $\boldsymbol{u} \sim \mN(\boldsymbol{0}, \boldsymbol{\Sigma}_{\mathrm{cr}})$, $\boldsymbol{v} \sim \mN(\boldsymbol{0}, \boldsymbol{\Sigma}_a)$ and $\boldsymbol{w} \sim \mN(\boldsymbol{0}, \boldsymbol{\Sigma}_a)$.
Then Eqn. (\ref{eq: rewrite}) can be rewritten as:
\begin{equation}
\label{eq: simple}
    \mmE_{\boldsymbol{u}, \boldsymbol{v}, \boldsymbol{w}} \left[ \boldsymbol{v} + \boldsymbol{w} \right]^ \top \left[ \boldsymbol{v} + \boldsymbol{w} \right] + \beta \left[\bP \boldsymbol{v}  {+} \boldsymbol{\mu}_\Delta {-} \boldsymbol{u}\right]^\top \left[\bP \boldsymbol{v}  {+} \boldsymbol{\mu}_\Delta {-} \boldsymbol{u}\right].
\end{equation}
Let $\boldsymbol{v}^* = \bP \boldsymbol{v}$, we obtain $\boldsymbol{v}^* \sim \mN(\boldsymbol{0}, \bP  \boldsymbol{\Sigma}_a \bP ^ \top)$.
Expanding Eqn. (\ref{eq: simple}) to:
\begin{equation}
\label{eq: expand}
\mmE_{\boldsymbol{u}, \boldsymbol{v}, \boldsymbol{w}} \left[ 
\boldsymbol{v}^\top\boldsymbol{v} + \boldsymbol{v}^\top\boldsymbol{w} + \boldsymbol{w}^\top\boldsymbol{v} +\boldsymbol{w}^\top\boldsymbol{w}
\right] {+}  \beta \left[ 
{\boldsymbol{v}^*}^\top\boldsymbol{v}^* {+}  \boldsymbol{\mu}_\Delta^\top \boldsymbol{\mu}_\Delta {+}  \boldsymbol{u}^\top\boldsymbol{u} +
{\boldsymbol{v}^*}^\top \boldsymbol{\mu}_\Delta {+}  \boldsymbol{\mu}_\Delta ^\top \boldsymbol{v}^* {-} {\boldsymbol{v}^*}^\top\boldsymbol{u} {-} \boldsymbol{u}^\top\boldsymbol{v}^* {-} \boldsymbol{\mu}_\Delta ^\top \boldsymbol{u} {-} \boldsymbol{u} ^\top \boldsymbol{\mu}_\Delta 
\right].
\end{equation}
Since $\boldsymbol{\mu}_\Delta$ is a constant, $\boldsymbol{u} \sim \mN(\boldsymbol{0}, \boldsymbol{\Sigma}_{\mathrm{cr}})$, $\boldsymbol{v} \sim \mN(\boldsymbol{0}, \boldsymbol{\Sigma}_a)$ and $\boldsymbol{v}^* \sim \mN(\boldsymbol{0}, \bP  \boldsymbol{\Sigma}_a \bP^ \top)$, we obtain $\mmE_{\boldsymbol{v}, \boldsymbol{w}}[\boldsymbol{v} ^\top \boldsymbol{w}] {=} \mmE_{\boldsymbol{v}, \boldsymbol{w}} [\boldsymbol{w} ^\top \boldsymbol{v}] =0$, $\mmE_{\boldsymbol{v}} [{\boldsymbol{v}^*}^\top \boldsymbol{\mu}_\Delta] = \mmE_{\boldsymbol{v}}  [\boldsymbol{\mu}_\Delta^\top {\boldsymbol{v}^*}]=0$ and $\mmE_{\boldsymbol{u}} [\boldsymbol{u}^\top \boldsymbol{\mu}_\Delta] = \mmE_{\boldsymbol{u}}  [\boldsymbol{\mu}_\Delta^\top \boldsymbol{u}]=0$.
Then, Eqn. (\ref{eq: expand}) can be represented as:
\begin{equation}
\label{eq: merge}
\mmE_{\boldsymbol{u}, \boldsymbol{v}, \boldsymbol{w}} \left[ 
\boldsymbol{v}^\top{\boldsymbol{v}}  + \boldsymbol{w}^\top\boldsymbol{w}
\right]+ \beta \left[
{\boldsymbol{v}^*}^\top\boldsymbol{v}^* + \boldsymbol{\mu}_\Delta^\top \boldsymbol{\mu}_\Delta + \boldsymbol{u}^\top\boldsymbol{u} - 2{\boldsymbol{v}^*}^\top\boldsymbol{u} 
\right].
\end{equation}
According to the property of trace of matrix, minimizing Eqn. (\ref{eq: merge}) is equivalent to minimizing:
\begin{align}
\label{eq: trance}
&tr \left(\mmE_{\boldsymbol{u}, \boldsymbol{v}, \boldsymbol{w}} \left[ 
\boldsymbol{v}\boldsymbol{v}^\top  + \boldsymbol{w} \boldsymbol{w}^\top
\right]+ \beta \left[
{\boldsymbol{v}^*}{\boldsymbol{v}^*}^\top + \boldsymbol{u}\boldsymbol{u}^\top - 2{\boldsymbol{v}^*}\boldsymbol{u} ^\top
\right] \right)\\
= & tr \left( 2 \boldsymbol{\Sigma}_a + \beta (\boldsymbol{\Sigma}_a +\boldsymbol{\Sigma}_{\mathrm{cr}} - 2 \mmE_{\boldsymbol{u}, \boldsymbol{v}} [{\boldsymbol{v}^*}\boldsymbol{u} ^\top])
\right)
\end{align}
Let $\Phi=\mmE_{\boldsymbol{u}, \boldsymbol{v}} [{\boldsymbol{v}^*}\boldsymbol{u} ^\top]$ denote the covariance of ${\boldsymbol{v}^*}$ and $\boldsymbol{u}$. Then, the optimal $T$ to Equation 6 can be reformulated as:
\begin{equation}
\label{eq: max trance}
    \bT=\underset{\bT}{\arg \max }({tr}(\Phi )).
\end{equation}
Olkin et al. \cite{olkin1982distance} show a unique solution to Eqn. (\ref{eq: max trance}) is 
\begin{equation}
\label{eq: Phi}
    \Phi = \bP \boldsymbol{\Sigma}_{a} \bP^\top\boldsymbol{\Sigma}_{\mathrm{cr}}^{1 / 2}\left(\boldsymbol{\Sigma}_{\mathrm{cr}}^{1 / 2} \bP  \boldsymbol{\Sigma}_{a} \bP ^ \top\boldsymbol{\Sigma}_{\mathrm{cr}}^{1 / 2}\right)^{-1 / 2} \boldsymbol{\Sigma}_{\mathrm{cr}}^{1 / 2}.
\end{equation}
Since $\Phi = \mmE_{\boldsymbol{u}, \boldsymbol{v}} [{\boldsymbol{v}^*}\boldsymbol{u} ^\top] = \mmE_{\boldsymbol{u}, \boldsymbol{v}} [{\boldsymbol{v}^*}(\bT^{-1} \boldsymbol{v}) ^\top]=\bP\mmE_{\boldsymbol{u}, \boldsymbol{v}} [\boldsymbol{v} \boldsymbol{v}^\top]  (\bT^{-1})^\top= \bP \boldsymbol{\Sigma}_{a}(\bT^{-1})^\top$, combining Eqn. (\ref{eq: Phi}) obtains the final $\bT$ \begin{equation}
    \bT = \bP ^{-1}\boldsymbol{\Sigma}_{\mathrm{cr}}^{-1 / 2}\left(\boldsymbol{\Sigma}_{\mathrm{cr}}^{1 / 2} \bP ^\top \boldsymbol{\Sigma}_{a} \bP  \boldsymbol{\Sigma}_{\mathrm{cr}}^{1 / 2}\right)^{1 / 2} \boldsymbol{\Sigma}_{\mathrm{cr}}^{-1 / 2 }.
\end{equation}
\end{proof}

\section{Inference of \sexyname}
\label{sec:sup_Inference}
\begin{figure*}[ht]
\centering
        \includegraphics[width=1\textwidth]{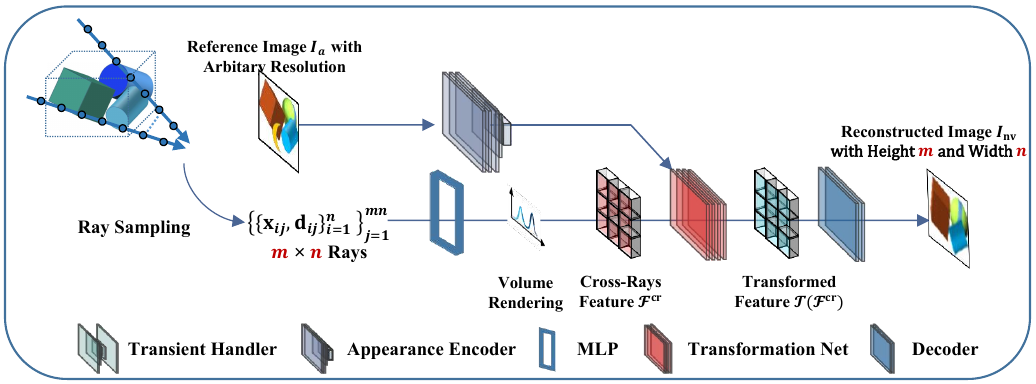}
    \caption{Illustration on inference of \sexyname.}
    \label{fig:sup_inference}
\end{figure*}
We provide details on inference of our \sexyname~in Fig.~\ref{fig:sup_inference} and Alg.~\ref{alg:crnerf_inference}. During inference, we sample $m \times n$ rays, which intersect with $m \times n$ pixels of reconstructed image $I_\mathrm{n}$ ($m$ and $n$ equals the height and width of $I_\mathrm{n}$). Thanks to our encoder parameterized by convolutional neural network and adaptive average pooling, the reference image can be of arbitrary size and we generate an appearance embedding $\mathcal{F}^\mathrm{a}$ by encoding $I_\mathrm{n}$ with appearance encoder. After representing the $m\times n$ rays with our proposed cross-ray feature $\mathcal{F}^\mathrm{cr}$, we fuse $\mathcal{F}^\mathrm{cr}$ and $\mathcal{F}^\mathrm{a}$ with a transformation net and then decode the fused feature to synthesize $I_\mathrm{n}$. During inference, we discard the transient object handler and content encoder.

\begin{algorithm}[t]
 \small 
	\caption{The Inference pipeline of \sexyname.}
	\label{alg:crnerf_inference}
	\KwIn{$m*n$ rays $\{\br_i\}_{i=1}^{m*n}$, a reference image $\mathcal{I}_{a}$ with size $m*n$, a multilayer perceptron $\mathrm{MLP}_{\theta_1}$, an appearance encoder $E_{\theta_2}$, a transformation net $\mT_{\theta_3}$, a decoder $D_{\theta_4}$.}
 \KwOut{The estimated colors of $m*n$ pixels of a novel view.} 
	
		Generate cross-ray features $\mF^{\mathrm{cr}}$
and appearance feature $\mF^{a}$ with $E_{\theta_2}$ and $\mathrm{MLP}_{\theta_1}$ by Eqn. (\ref{eq:style_ours_generating}). 

Injecting appearance from $\mathcal{I}_{a}$ to scene representation by fusing $\mF^{\mathrm{cr}}$ and $\mF^{a}$ via $\mT_{\theta_3}$. 

Estimating color $\hat{\bc}(\{\br_i\}_{i=1}^{m*n})$ \wrt~the rays and the reference image by leveraging $D_{\theta_4}$.

\end{algorithm}

  \begin{minipage}{\textwidth}
  \begin{minipage}[c]{0.49\textwidth}
    \centering
    \includegraphics[width=0.9\textwidth]{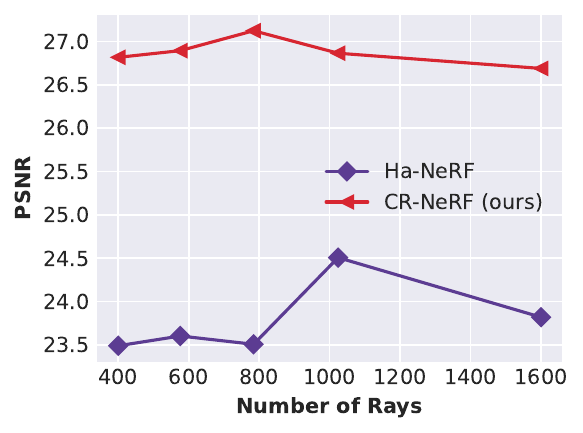}
    \captionof{figure}{Effectiveness of different number of rays on our CR-NeRF and Ha-NeRF on brandenburg dataset in terms of PSNR.}
 \label{fig:numberOfRay}
  \end{minipage}
  \hfill
  \begin{minipage}[c]{0.49\textwidth}
    \centering
\begin{tabular}{cccccc}
\toprule
    \#Rays   & 400 & 576 & 784 & 1024 & 1600 \\
        \midrule
Ha-NeRF & 23.49                 & 23.60                 & 23.51                 & 24.51                  & 23.82   \\
CR-NeRF & 26.82                 & 26.89                  & 27.12                 & 26.86                  & 26.69                  \\
\bottomrule
\end{tabular}
      \captionof{table}{Effectiveness of different number of rays on our CR-NeRF and Ha-NeRF on brandenburg dataset in terms of PSNR.}
    \end{minipage}
  \end{minipage}

\section{Effectiveness of Number of Rays}
\label{sec:effectofrays}

We analyze the impact of the number of rays (\#rays) on both Ha-NeRF and our (\sexyname). Fig.~\ref{fig:numberOfRay} shows the PSNR results of the two methods on the Brandenburg dataset in terms of different values of \#rays. \sexyname~consistently outperforms Ha-NeRF across all tested values of \#rays, which verifies that considering multiple rays consistently boosts the performance of \sexyname. Additionally, the performance of \sexyname~increases as the number of rays increases. However, we also note that after the number of rays exceeds 784, the performance of \sexyname~starts to degrade gradually. One possible explanation is that increasing \#rays over a threshold introduces ambiguity in view-consistent modeling, which harms the quality of synthesized views. Note that although Ha-NeRF uses multiple rays as input, information from each individual ray does not intersect with that of the others.

\section{Effectiveness of Cross-Ray Paradigm and Fusing Level}
\label{sec:sup_Cross-Ray}
\begin{figure*}[ht]
    \centering
    \includegraphics[width=1\textwidth]{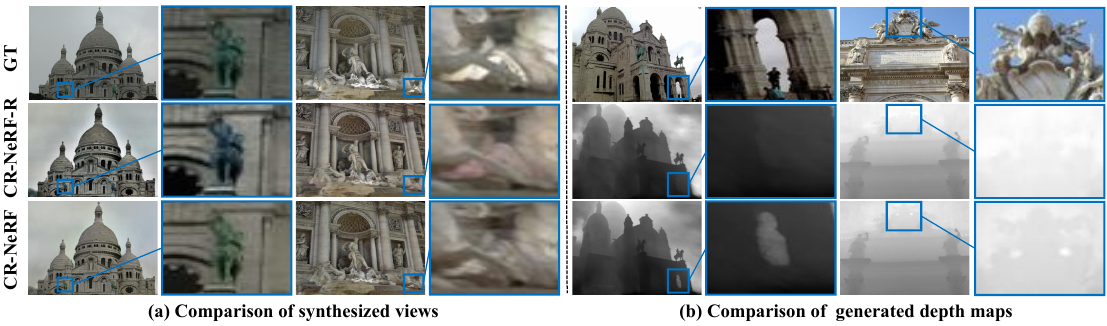}
    \caption{Comparison of \sexyname~and \sexyname-R regarding detailed appearance and depth maps. \sexyname~is able to synthesize a more accurate appearance (\eg~the color of the statue in Trevi Fountain and in Sacre Coeur). Moreover, \sexyname~ successfully estimates the depth of the cavity portion of the building while \sexyname-R fails.}
    \label{fig:baseline_img}
\end{figure*}

We study the effectiveness of our cross-ray paradigm and on which level to fuse with appearance features. To this end, we construct \sexyname-R, the only difference of \sexyname-R and \sexyname~is that \sexyname-R conduct appearance transfer by considering \textit{ray points of different rays} but \sexyname~achieves the transfer on \textit{different rays}.
In other words, \sexyname-R fuses an \textit{image-level} appearance feature $\mathcal{F}^{a}$ with \textit{ray-point level} features, while \sexyname~combines $\mathcal{F}^{a}$ and $\mathcal{F}^{\mathrm{cr}}$.
From Fig.~\ref{fig:baseline_img}, \sexyname~is able to model a more accurate appearance, while also reconstructing a more consistent geometry. These results verify the superiority of the cross-ray manner and show fusing the image-level appearance features with cross-ray features is more effective than with the cross-ray-points features.

\section{Interpolation of Appearance Embedding}
\label{sec:sup_Interpolation}

\begin{figure*}[ht]
    \centering
\includegraphics[width=0.82\textwidth]{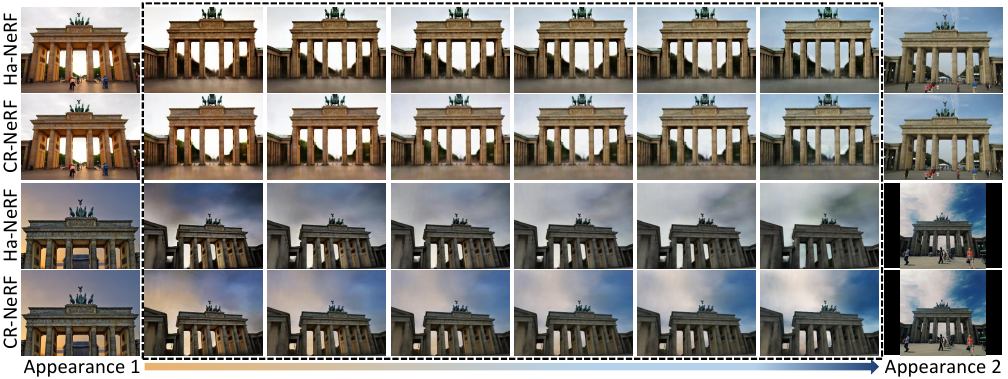}
    \caption{Interpolating between appearance 1 and appearance 2 with a fixed camera position (synthesized results are in the dashed box).}
    \label{fig:interpolation_sup}
\end{figure*}

Our proposed \sexyname~is able to synthesize images that gradually change from one appearance image to another. We achieve this by linearly interpolating the appearance features of the two appearance images. From Fig.~\ref{fig:interpolation_sup}, we observe that (1) \sexyname~ is able to handle transient objects and thus synthesize non-transient images (\eg~images in the second row of Fig.~\ref{fig:interpolation_sup} have no transient objects, such as visitors in appearance 1, and the ground synthesized by \sexyname~better shows the reflection effect of ground in appearance 1). (2) \sexyname~captures the appearance more accurately than Ha-NeRF (\eg~the sky color in the third row of is not as accurate as the fourth row of Fig.~\ref{fig:interpolation_sup}).

\section{Modeling Appearance from Brandenburg and Trevi}
\label{sec:sup_Appearance}
\begin{figure*}[ht]
    \centering
\includegraphics[width=0.85\textwidth]{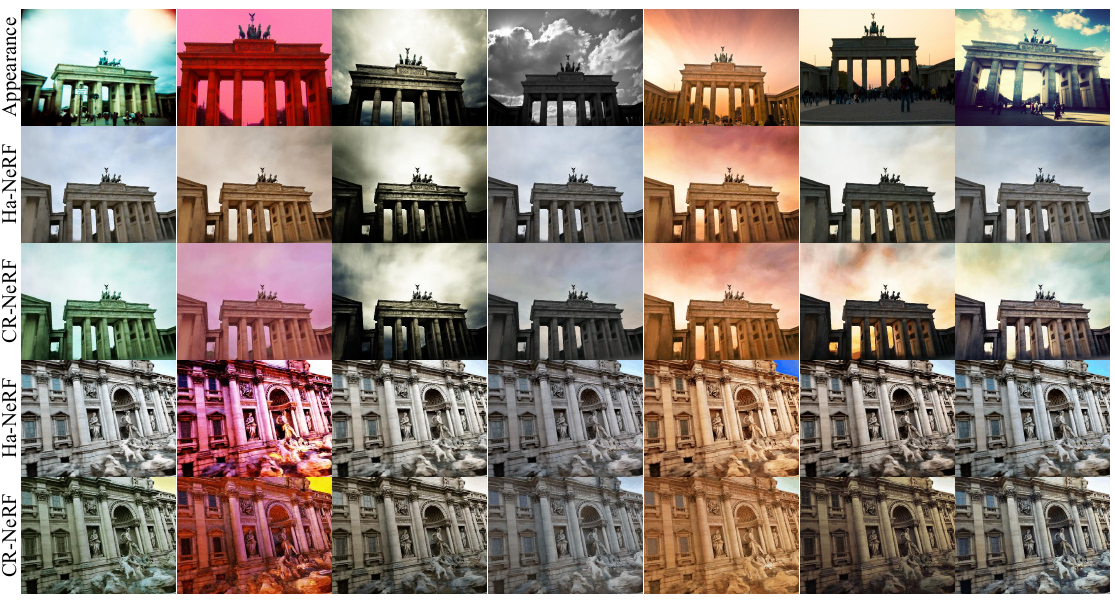}
    \caption{Transferring appearance from Brandenburg Gate to Brandenburg Gate and Trevi Fountain.}
    \label{fig:branden_style}
\end{figure*}

\begin{figure*}[ht]
    \centering
\includegraphics[width=0.8\textwidth]{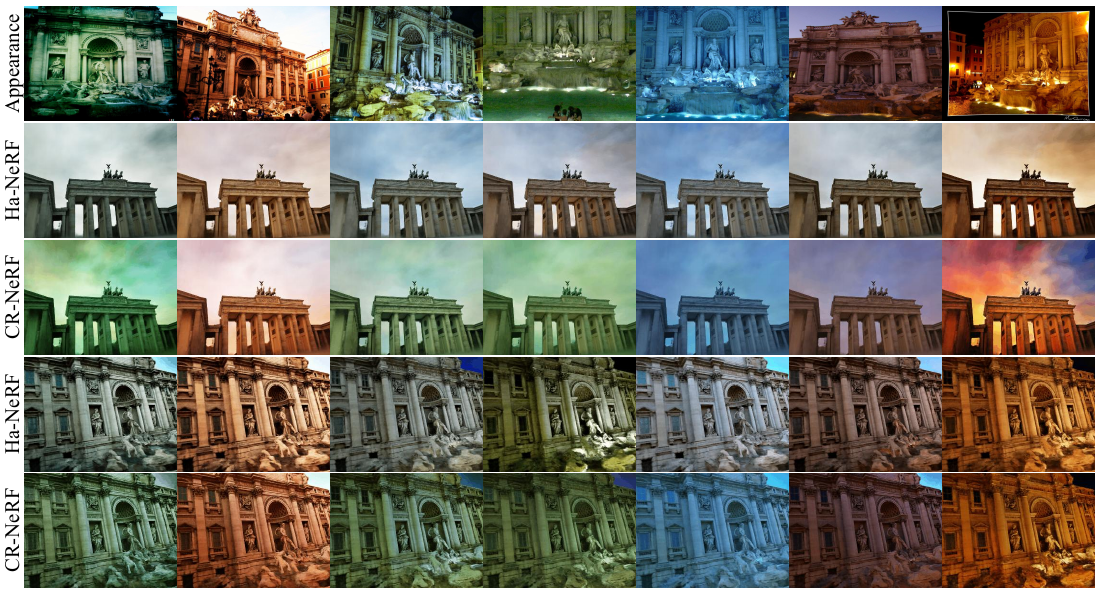}
    \caption{Transferring appearance from Trevi Fountain to Brandenburg Gate and Trevi Fountain.}
    \label{fig:trevi_style}
\end{figure*}

We show qualitative experimental results of appearance modeling using images from Brandenburg and Trevi. As shown in Fig.~\ref{fig:branden_style} and Fig.~\ref{fig:trevi_style}, we transfer appearance from Brandenburg to Brandenburg and Trevi and vice versa.  \sexyname~recovers a more accurate appearance than Ha-NeRF, which demonstrates the effectiveness of our cross-ray paradigm.

\section{Modeling Appearance from Unseen Images}
\label{sec:sup_Unseen}
\begin{figure*}[ht]
    \centering
\includegraphics[width=0.8\textwidth]{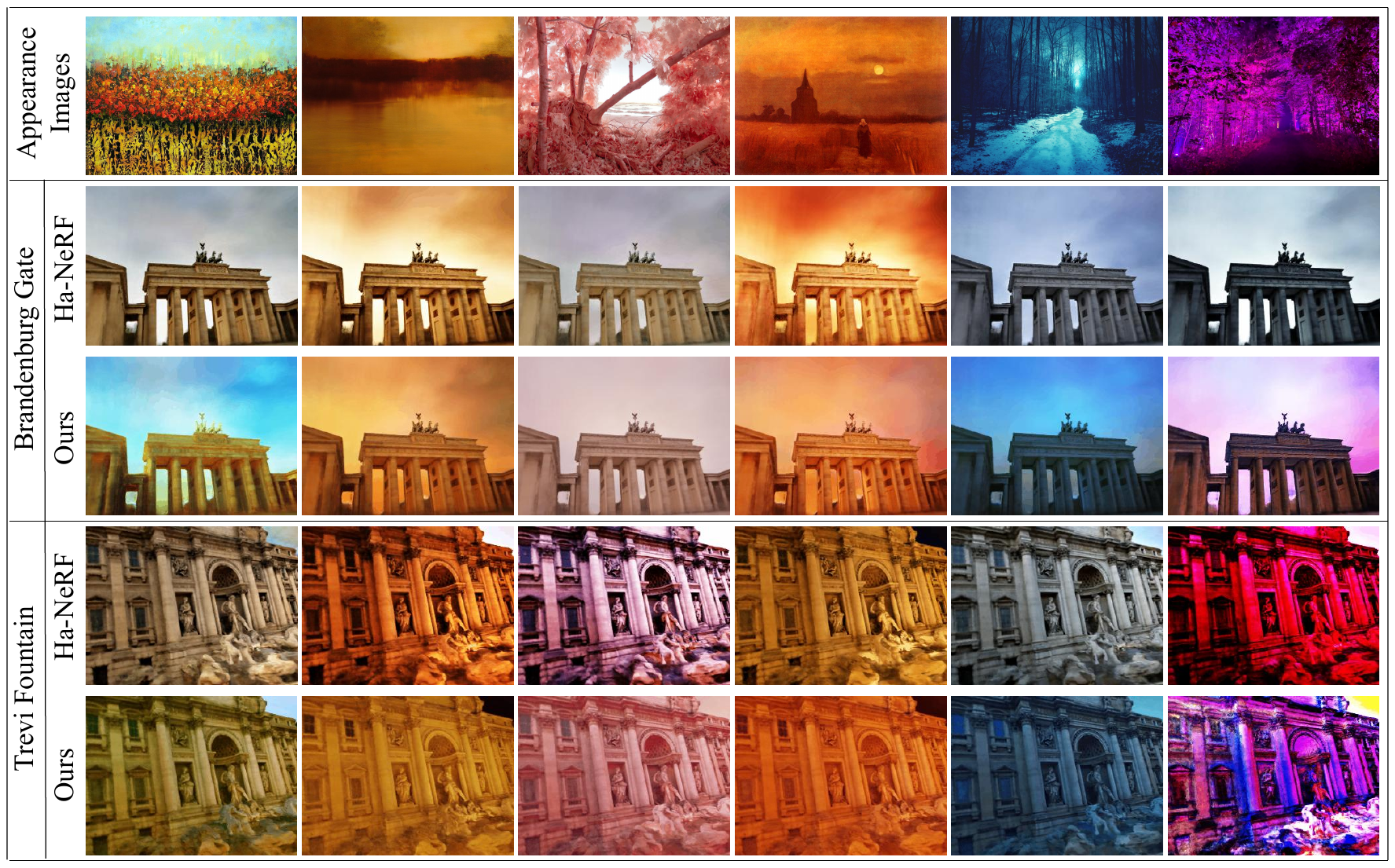}
    \caption{Transferring appearance from unseen images to Brandenburg Gate and Trevi Fountain datasets.}
    \label{fig:unseen_style}
\end{figure*}

Our proposed \sexyname~is able to deal with unseen appearance images thanks to the ability of our cross-ray appearance modeling handler.  As shown in Fig.~\ref{fig:unseen_style}, our \sexyname~captures the whole range appearance (\eg~the blue and purple appearance in the last two columns in Brandenburg and Trevi fountain datasets) of the reference image more accurately compared with Ha-NeRF. Moreover,  our \sexyname~synthesizes a more consistent appearance than images generated by Ha-NeRF (\eg~the sudden bright light on the sky of the second and fourth column in the Brandenburg dataset). We also provide videos of unseen transfers on videos in the supplementary material. Note that NeRF-W needs to optimize its appearance embedding on each test image by pixel-level supervision, thus NeRF-W cannot be directly applied to unseen appearance transfer. 

\section{Grid Sampling strategy}
\label{sec:grid_sample}
Grid sampling strategy aims to extract a grid of $k \times k$ image pixels from a reference image, guided by a grid center $\mathbf{u}$ and sampling scale $s$.  As detailed in [36] and illustrated in Fig.~\ref{fig:grid}, GS involves uniformly selecting $k \times k$ image pixels based on the coordinate set $\mathcal{P}(\mathbf{u}, s)=\left\{(s x+u_x, s y+u_y) \mid x, y \in\left\{-\frac{k}{2}, \ldots, \frac{k}{2}-1\right\}\right\}$, where $\mathbf{u}{=}(u_x,u_y) {\in} \mathbb{R}^{2}$ and $s {\in} \mathbb{R}^{+}$. With these pixel coordinates, we sample $k \times k$ rays for cross-ray synthesis and also sample our predicted visibility mask for transient handling.

\section{Comparison of Training Time}
\label{sec:training_time}
In Tab.~\ref{tab:trainingtime}, we report the training times for \sexyname, Ha-NeRF, and NeRF-W, spanning 20 epochs, are 1583, 1701, and 1504 minutes, respectively. We employ 8 TITAN Xp GPUs with 17200 iterations per epoch.

\section{Comparisons of Transient Network}
\label{sec:compare_transient}
To further verify the effectiveness of our proposed transient network, we conduct a comparative analysis by replacing the transient network in our \sexyname~with that of NeRF-W (termed \sexyname-U) and utilizing uncertainty formulation for training. The results in Tab.~\ref{tab:ablation_uncertainty} show the superiority of our transient network on three datasets. Moreover, we visualize the output of the transient networks of \sexyname~and NeRF-W in Fig.~\ref{fig:supp_transient}. \sexyname~achieves a more accurate prediction by identifying the semantic feature of tourists and trees. Our transient network outperforms NeRF-W because predicting object visibility is much easier than predicting the colors and densities of transient objects.



\begin{minipage}{\textwidth}
\begin{minipage}[c]{0.49\textwidth}
    \centering
       \includegraphics[height=0.350\linewidth]{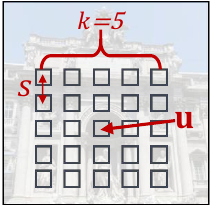}
   \captionof{figure}{Illustration of grid sampling strategy.}
   \label{fig:grid}
    \end{minipage}
  \begin{minipage}[c]{0.5\textwidth}
    \centering
    \includegraphics{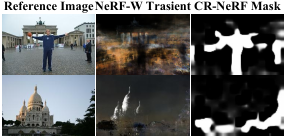}
    \captionof{figure}{Comparisons of the transient networks of \sexyname~and NeRF-W.}
    \label{fig:supp_transient}
\end{minipage}
  \end{minipage}

\begin{minipage}{\textwidth}
  \begin{minipage}[c]{0.49\textwidth}
    \centering
    \begin{tabular}{lccc}
\toprule
&Brandenburg &Sacre &Trevi\\
\midrule
\sexyname-U~            & 24.72/0.8873         & 20.88/0.8161      & 20.84/0.7382       \\  
\sexyname~              & \textbf{26.86/0.9069}       & \textbf{22.03/0.8369}    & \textbf{22.02/0.7488} \\
\bottomrule
\end{tabular}
 \captionof{table}{PSNR/SSIM of \sexyname-U on three datasets.}
 \label{tab:ablation_uncertainty}
  \end{minipage}
  \hfill
  \begin{minipage}[c]{0.49\textwidth}
    \centering
    \begin{tabular}{cccc}
    \toprule
        Method & EPOCH & Iteration & Time (minutes) \\ 
        \midrule
        NeRF-W & 20 & 17200 & 1504 \\ 
        Ha-NeRF & 20 & 17200 & 1701 \\ 
        CR-NeRF & 20 & 17200 & 1583 \\ 
    \bottomrule
    \end{tabular}
      \captionof{table}{Training time comparisons of different methods.}
      \label{tab:trainingtime}
    \end{minipage}
  \end{minipage}


\end{document}